\documentclass[10pt]{article} % For LaTeX2e
\usepackage[accepted]{rlc}
\usepackage{amsmath}
\usepackage{amssymb}            % Defines common symbols like \mathbb R
\usepackage{mathtools}          % Extends amsmath, providing common math tools
\usepackage{graphicx}           % For including images
\usepackage{subcaption}         % Allows for the use of subfigures and subcaptions
\usepackage[space]{grffile}     % For spaces in image names
\usepackage{url}                % For displaying URLs
\usepackage[bb=boondox]{mathalfa}
\usepackage{xspace}

\usepackage{microtype}
\usepackage{graphicx}
\usepackage{booktabs} % for professional tables
\usepackage{algorithm}
\usepackage{algorithmicx}
\usepackage{algpseudocode}
\algnewcommand{\algorithmicforeach}{\textbf{for each}}
\algdef{SE}[FOR]{ForEach}{EndForEach}[1]
  {\algorithmicforeach\ #1\ \algorithmicdo}% \ForEach{#1}
  {\algorithmicend\ \algorithmicforeach}% \EndForEach
\newcommand{\pushright}[1]{\ifmeasuring@#1\else\omit\hfill$\displaystyle#1$\fi\ignorespaces}
\algrenewcommand\algorithmicindent{0.8em}%
\usepackage{subcaption}
\usepackage{tikz}
\usepackage{amsmath}
\usepackage{wrapfig}
\usepackage{cleveref}

\usetikzlibrary{tikzmark}
\usetikzlibrary{calc}
\newcommand{\ALGtikzmarkcolor}{black}% customise this, if you want
\newcommand{\ALGtikzmarkextraindent}{2pt}% customise this, if you want
\newcommand{\ALGtikzmarkverticaloffsetstart}{-.5ex}% customise this, if you want
\newcommand{\ALGtikzmarkverticaloffsetend}{-.5ex}% customise this, if you want
\makeatletter
\newcounter{ALG@tikzmark@tempcnta}

\newcommand\ALG@tikzmark@start{%
    \global\let\ALG@tikzmark@last\ALG@tikzmark@starttext%
    \expandafter\edef\csname ALG@tikzmark@\theALG@nested\endcsname{\theALG@tikzmark@tempcnta}%
    \tikzmark{ALG@tikzmark@start@\csname ALG@tikzmark@\theALG@nested\endcsname}%
    \addtocounter{ALG@tikzmark@tempcnta}{1}%
}

\def\ALG@tikzmark@starttext{start}
\newcommand\ALG@tikzmark@end{%
    \ifx\ALG@tikzmark@last\ALG@tikzmark@starttext
    \else
        \tikzmark{ALG@tikzmark@end@\csname ALG@tikzmark@\theALG@nested\endcsname}%
        \tikz[overlay,remember picture] \draw[\ALGtikzmarkcolor] let \p{S}=($(pic cs:ALG@tikzmark@start@\csname ALG@tikzmark@\theALG@nested\endcsname)+(\ALGtikzmarkextraindent,\ALGtikzmarkverticaloffsetstart)$), \p{E}=($(pic cs:ALG@tikzmark@end@\csname ALG@tikzmark@\theALG@nested\endcsname)+(\ALGtikzmarkextraindent,\ALGtikzmarkverticaloffsetend)$) in (\x{S},\y{S})--(\x{S},\y{E});%
    \fi
    \gdef\ALG@tikzmark@last{end}%
}
\usepackage[english]{babel}
\title{Surprise-Adaptive Intrinsic Motivation for \\
Unsupervised Reinforcement Learning}

\author{%
   Adriana Hugessen\thanks{Equal contribution.}\\
   \And
   Roger Creus Castanyer\footnotemark[1]\\
   \And
   Faisal Mohamed\footnotemark[1]\\
   \And
   Glen Berseth \\
   \And \vspace{-2.3em} \\
     Université de Montréal and Mila Quebec AI Institute
     \\
 \texttt{\{adriana.knatchbull-hugessen,roger.creus-castanyer,faisal.mohamed\}@mila.quebec}
 }

\begin{document}

\maketitle

\begin{abstract}
    Both entropy-minimizing and entropy-maximizing objectives for unsupervised reinforcement learning (RL) have been shown to be effective in different environments, depending on the environment's level of natural entropy. However, neither method alone results in an agent that will consistently learn intelligent behavior across environments. In an effort to find a single entropy-based method that will encourage emergent behaviors in any environment, we propose an agent that can adapt its objective online, depending on the entropy conditions it faces in the environment, by framing the choice as a multi-armed bandit problem. We devise a novel intrinsic feedback signal for the bandit, which captures the agent's ability to control the entropy in its environment. We demonstrate that such agents can learn to optimize task returns through entropy control alone in didactic environments for both high- and low-entropy regimes and learn skillful behaviors in certain benchmark tasks. Videos and summarized findings can be found on our \href{https://sites.google.com/view/surprise-adaptive-agents?usp=sharing}{project webpage}\footnote{Code is available at \url{https://github.com/roger-creus/surprise-adaptive-agents}}.
\end{abstract}

%%%%%%%%%%%%%%%%%%%%%%%%%%%%%%%%%%%%%%%%%%%%%%%%%%%%%%%%%%%%%%%%
%% Section: Submission of papers to RLC
%%%%%%%%%%%%%%%%%%%%%%%%%%%%%%%%%%%%%%%%%%%%%%%%%%%%%%%%%%%%%%%%
\section{ Introduction}

Unsupervised reinforcement learning (URL), or learning without access to an extrinsic reward function, has recently gained significant attention, often as a pretraining method~\citep{jaderberg2017reinforcement} or as a reward bonus in sparse reward domains~\citep{schmidhuber1991curious,pathak2017curiosity,burda2018exploration}.
A recent focus has been on developing objectives where the agent has no access to extrinsic rewards and instead develops emergent behaviors from intrinsic motivation alone ~\citep{lopes2012exploration,kim2020active,berseth2019smirl}. In this context, unsupervised RL holds the promise of being able to develop natural-like intelligence, i.e. generally-capable agents that can be deployed to solve diverse tasks across diverse environments. However, thus far, no single intrinsic motivation function has succeeded in capturing the complexity of motivation that gives rise to intelligent systems.

Interestingly, two seemingly opposing methods, surprise-minimization \citep{friston2010free,berseth2019smirl} and surprise-maximization \citep{schmidhuber1991curious,pathak2017curiosity,hazan2019provably,tiapkin2023fast}, have been proposed as intrinsic motivations, with both methods performing well depending on the properties of the environment in which they are deployed. In general, surprise-minimizing methods \citep{berseth2019smirl} perform well in environments with naturally high entropy that can be reduced through control, while curiosity-based methods \citep{pathak2017curiosity} are better suited to environments where explicit exploration is necessary to encounter novel information. However, both methods are known to possess failure modes when exposed to the opposite entropy regime \citep{schmidhuber2010formal, sun2020dark}. 

In this work, we propose an adaptive mechanism to select between maximizing and minimizing surprise in a given environment, based on the agent's ability to exert control over its entropy conditions, which we frame as a multi-armed bandit problem.
We experimentally validate our \textit{surprise-adaptive} agent by demonstrating its ability to mirror a surprise-maximizing or -minimizing agent in didactic low- and high-entropy environments, respectively, and, in doing so, perform well on these tasks without any access to extrinsic task rewards. In benchmark environments, we demonstrate more diverse emergent behaviors, as measured by the performance on extrinsic task reward, than observed from the single-objective agents.
%%GB.02.29.24: We need to spice up the description of our experiments to sound more interesting. Describe the results on the Atari environments.

\section{Related work} \label{sec:relatedwork}
There is a rich body of work in the field of unsupervised RL and intrinsic motivation, upon which our method builds. The most widely explored class of intrinsic objectives is related to improving state coverage through exploration bonuses which reward some measure of novelty. In low-dimensional settings, count-based methods \citep{auer2002using, bellemare2016unifying,machado2020count} are simple and effective but do not always extend well to higher dimensions \citep{lobel2023flipping}. Another popular class of methods in high-dimensional settings uses prediction error as an exploration bonus \citep{schmidhuber1991curious,pathak2017curiosity}. A conceptually similar idea is that of entropy maximization \citep{hazan2019provably,tiapkin2023fast,jain2024maximum}, which seeks to maximize the entropy of the distribution of states experienced by the agent throughout its lifetime. Naive implementations of these novelty-seeking agents, however, can be susceptible to what is known as the "noisy-TV problem"  \citep{schmidhuber2010formal}, where the agent becomes transfixed by irreducible aleatoric noise in the environment. Various formulations have been developed to combat this issue, though issues often persist \citep{houthooft2016vime,pathak2017curiosity,burda2018exploration}. Though these methods are generally implemented as bonuses to the extrinsic reward, some works have also investigated the ability of curiosity-driven agents to achieve good task rewards without any access to the extrinsic reward \citep{burda2018large}

An alternative class of intrinsic objectives also targets the scenario where no extrinsic rewards are available by incentivizing the agent to exert control over its environment. This class of methods is rooted in the free energy principle, a concept from neuroscience that posits that intelligent organisms seek out stable niches by learning to control their environment to minimize the entropy they experience over their lifetime \citep{friston2010free}. One prominent formulation in this class is that of empowerment, defined as the maximal mutual information between an agent's actions and future states \citep{klyubin2005empowerment,karl2015efficient,zhao2020efficient}. However, empowerment is computationally difficult to estimate in large or continuous state and action spaces. A more tractable approximation to the free-energy principle was proposed by \cite{berseth2019smirl} as surprise-minimization. In this formulation, an upper bound on an agent's total trajectory entropy is minimized by rewarding the agent with the log-probability of the current state under the estimated state marginal distribution. This method has shown promising results in a diverse set of stochastic environments \citep{berseth2019smirl, rhinehart2021information}. Surprise-minimizing agents, however, can fall victim to the "dark room problem" \citep{sun2020dark}, where the agent discovers an area of the state-space without any stochastic dynamics and fails to seek out any additional experience.

Two recent works make efforts towards combining surprise-minimization and maximization objectives to avoid the degenerate cases of prior methods, either using a complex multi-agent paradigm~\citep{fickinger2021explore} or learned skills~\citep{zhao2022mixture}. In \cite{fickinger2021explore} the authors seek to capture more complex behaviors by alternately minimizing and maximizing surprise in an adversarial game between surprise-minimizing and surprise-maximizing players. However, this adversarial approach is susceptible to unstable training dynamics. In \cite{zhao2022mixture}, they circumvent the complexity of adversarial RL, instead training a single agent equipped with two different skill sets, one surprise-minimizing, and the other surprise-maximizing. This approach is conceptually and practically simpler than the multi-agent approach. However, neither method uses an adaptive mechanism to control the objective, instead using fixed-length windows to alternate between objectives. In contrast, our proposed method can adapt to entropy conditions online to bias the agent towards the objective with the greatest potential.
%%GB.02.29.24: The above should be shortened. This is too long.

Prior works have explored adaptivity in RL and found that it can be beneficial for learning \citep{badia2020agent57, moskovitz2021tactical}. Similar to our work, \citet{moskovitz2021tactical} uses a multi-armed bandit to control a learning hyperparameter. However, their method relies on extrinsic rewards for providing feedback to the bandit, while our method derives rewards based only on intrinsic signals.
%%GB.02.29.24: The above might be too much detail for the related work section. However, adding references on how adaptability is helpful should be added to the introduction to further motivate the method design.

\section{Background} \label{maths}

\paragraph{Reinforcement learning.}
RL is a learning paradigm for sequential decision-making problems. In RL, an agent acts in an environment from which it receives observations and rewards. Formally, this process can be modelled as a Markov Decision Process (MDP) consisting of the tuple $(\mathcal{S}, \mathcal{A}, \mathcal{T}, \mathcal{R}, \gamma)$ where $\mathcal{S}$ is the state space, $\mathcal{A}$ is the action space, $\mathcal{T}: \mathcal{S} \times \mathcal{A} \times
\mathcal{S} \rightarrow [0, 1]$ is the transition function, $\mathcal{R} : \mathcal{S} \times \mathcal{A} \rightarrow \mathcal{R}$ is the reward function, and $\gamma$ is the discount factor. The goal of the RL agent is to find a policy $\pi_{\phi}$ that produces actions that maximize the expected sum of discounted future rewards. 
\begin{equation}
     \pi_{\phi}(a_t|s_t) = \text{argmax}_{\phi}\mathbf{E}_{p(\tau|\phi)}\left[\sum_{t=0}^T\gamma^tr(s_t,a_t)\right]
      \label{eq:rl_obj}
\end{equation}
In our experiments, we use the value-based method DQN \citep{mnih2015playing} to solve \autoref{eq:rl_obj}. 
%%GB.02.29.24: DO we only use DQN?

\paragraph{Multi-armed bandits}
Multi-armed bandits can be thought of as a special case of RL where the state-space consists only of a single state. Typically evaluated based on regret, multi-armed bandit algorithms focus on the efficient trade-off between exploration and exploitation in order to find an optimal action while incurring the minimum amount of sub-optimal actions. In this work, we adopt one of the most popular algorithms, Upper Confidence Bounding (UCB)\citep{lai1985asymptotically} for an efficient trade-off. The UCB algorithm adds a count-based exploration bonus to the current value estimate of an action before selecting the maximum valued arm:
\begin{equation}
    a_t = \text{argmax}_{a \in \mathcal{A}}\left(Q_t(a) + \sqrt{\frac{\log{t}}{N_t(a)}}\right)
\end{equation}
\paragraph{Entropy and surprise}
The notion of surprise derives from the optimization of the entropy of the state marginal distribution under the policy $\pi_{\phi}(a|s)$, which we denote  $d^{\pi_{\phi}}(s_t)$. Given an estimate of this state marginal distribution, $p_{\theta_{t-1}}(s_t)$, we can express an estimate of the sum of the entropies of the state distribution across a trajectory (see Appendix A of \citet{berseth2019smirl} for a full derivation):
\begin{align}
\sum_{t=0}^T \mathcal{H}(s_t) &= \sum_{t=0}^T - \mathbf{E}_{s_t \sim d^{\pi_{\phi}}(s_t)}\left[\log d^{\pi_{\phi}}(s_t)\right] \leq \sum_{t=0}^T \mathbf{E}_{s_t \sim d^{\pi_{\phi}}(s_t)}\left[-\log p_{\theta_{t-1}}(s_{t})\right]
\label{eq:entropy}
\end{align}
Recalling Equation \ref{eq:rl_obj}, we can see that minimizing the sum of the state entropy over a trajectory (\autoref{eq:entropy}) corresponds to a surprise minimizing agent \citep{berseth2019smirl} with a reward function given by:
\begin{equation}
    \label{eq:smirl_reward}
    r_{\text{s-min}}(s_t, a_t) = \log p_{\theta_{t}}(s_{t+1})
\end{equation}
Maximizing this objective corresponds to an RL agent with a reward function given by: 
\begin{equation}
    \label{eq:negative_smirl_reward}
    r_{\text{s-max}}(s_t, a_t) = -\log p_{\theta_{t}}(s_{t+1})
\end{equation}
which is similar to the rewards provided to the EntGame agent in \citet{tiapkin2023fast}. 

Conceptually, this means that the agent is punished (or rewarded) if the observed state $s_t$ is "surprising", that is, if it has high negative log-likelihood under the state marginal distribution estimated so far. Hence, we refer to \Cref{eq:smirl_reward} as surprise-minimization and \Cref{eq:negative_smirl_reward} as surprise-maximization.

\section{Surprise-adaptive bandit} \label{method}

%%GB.02.29.24: Is there some kind of figure we can create to understand the method. We could visualize our estimates for the entropy calculations.

%%GB.02.29.24: This section should follow the story (1) first we need to estimate entropy of the marginal state distribution. Introduce the different ways this can be done. You all have explored many options for different types of distributions (we should have an ablation for different versions). (2) Now that we can estimate the entropy well, we need a mechanism to switch the agent between min vs max. Now introduce the details on the bandit and how it works, and the RL algorithm. Also, talk about how the policy is conditioned on an indicator for the two options for the bandit.

Surprise-minimization and surprise-maximization are most effective under particular entropy conditions in the environment, surprise-minimization under high-entropy conditions~\citep{berseth2019smirl}, and surprise-maximization under low-entropy conditions. An intrinsically motivated agent that could capitalize on the advantages each objective provides in its respective entropy regime would be a more powerful and versatile intrinsic learner. Hence, we propose an agent that can alternatively optimize for either objective, with an online adaptive mechanism for selecting the objective.

To design such an adaptive agent, we must first be able to optimize for either single-objective, which requires an estimation of the state marginal distribution at time $t$, parameterized by $\theta_t$ (denoted $p_{\theta_{t-1}}$ in \Cref{eq:entropy}). In general, this estimation can be quite complex; In \citet{berseth2019smirl}, the authors propose a simplification which we adopt here. The method estimates $\theta_t$ by first selecting an appropriate distribution family to model observations (i.e. Gaussian, Bernoulli, etc.) and using maximum likelihood estimation to estimate the sufficient statistics of the distributions, fitted to the history of observed states through time $t$. Further details on estimating the state marginal distribution as well as ablations on the choice of distribution are provided in \Cref{sec:est_state_marg}.

% To create an agent that can adjust its objective depends on its ability to estimate the entropy of the policies stationary distirubiton. In general, this estimation is complex and we explore a few options for how to best estimate the entropy of the environment and the agent's policy. 
% For example, how can we best capture the entropy distribution for image-based environments where simple Gaussian or Bernoulli distributions are ineffective? These choices are important because the agents' performance and ability to choose between the best entropy control objectives depend on how well we can estimate the stationary distribution.
% We find that for image-based environments, it is best to use a combination of Bernoulli split across channels~\todo{ref results section}.
% The method for estimating the sufficient statistics of the state marginal distribution is discussed in more detail in section \ref{sec:results}.

To adaptively select between the two objectives online, we propose a multi-armed bandit approach. Provided with an appropriate performance signal, a bandit learns to select optimally between actions, trading off exploration with exploitation to minimize the overall regret, making it an appropriate choice for online adaption. The key question is what type of feedback is best to provide the bandit, given access only to intrinsic rewards. We propose a feedback mechanism grounded in the observation that the general goal in both surprise minimization and surprise maximization is for the agent to be able to affect a change in the level of surprise it experiences. In a low-entropy environment, the agent can best affect change by increasing entropy, and vice versa.  

We propose using the absolute percent difference between the entropy of the state marginal distribution at the end of the $m$th episode  ($H(p_{\theta_T}^{(m)})$) 
%%GB.02.29.24: What does the T stand for? Does this indicate the calculation depends on the length of the trajectory? That is not stated anywhere.
and that of a \textit{random agent} in the same environment ($H(p_{\theta_T}^{\text{rand}})$) (\Cref{eq:feedback}). The motivation for this is as follows: a random agent cannot control the environment entropy as it cannot take any actions in response to feedback. Agents that produce state visitation distributions whose entropy significantly diverges from that of a random agent must therefore be exerting control over the entropy in the environment. We therefore provide feedback to the bandit which promotes agents that can exert such control by rewarding large deviations from a random agent. Since we are approximating the state marginal distributions by an analytical distribution, we can compute $H(p_{\theta_T}^{(m)})$ analytically from the estimated parameters (see \ref{sec:est_state_marg} for further details).
\begin{equation}
\label{eq:feedback}
    f_m = \left|\frac{H(p^{(m)}_{\theta_{T}}) - H(p_{\theta_T}^{\text{rand}})}{H(p_{\theta_T}^{\text{rand}})}\right|
\end{equation}
The precise algorithm is as follows (\Cref{alg:surprise-adapt}). At the start of training, we estimate the entropy of a random agent by collecting trajectories using a uniform random policy and averaging the entropy of the state marginal distributions, computed at the end of each trajectory (\Cref{alg:rand}). Then, at the start of each episode $m$, we select an arm from the bandit, represented by binary indicator $\alpha^{(m)}$, according to the UCB algorithm (\Cref{alg:select_bandit}),  which determines if the agent will receive rewards according to \autoref{eq:smirl_reward} or \autoref{eq:negative_smirl_reward}  during the upcoming episode. The agent is trained for a single episode, using any RL algorithm (\Cref{alg:rl}). At the end of each episode, the bandit receives feedback $f_m$ on its selection (\Cref{alg:feedback}). Algorithm \ref{alg:surprise-adapt} shows the full training procedure.

To instantiate the surprise-adaptive agent, we construct an augmented MDP out of the original Markov process. Following \citet{berseth2019smirl}, this augmented MDP has a state space that includes the original state $s_t$, as well as the sufficient statistics of the state marginal distribution $\theta_{t}$. We additionally include $\alpha^{(m)}$, as defined above, to ensure the reward function remains Markovian \citep{castanyer2023improving}. 

In our experiments, all agents were trained using DQN \citep{mnih2015playing}, using two convolutional neural networks (CNN) to encode the state.
The first CNN encodes the observed state $s_t$, while the second encodes the sufficient statistic of the state marginal distribution $\theta_t$ along with the bandit choice $\alpha^{(m)}$ which is added as an additional channel before applying the CNN. The outputs of the CNNs are concatenated and passed through an MLP that outputs the Q-value.  More details on environments and training can be found in \Cref{appendix}. 
%%GB.03.05.24: This should be in the methods section. It is not related to the experimnets.

\begin{algorithm}[tbh]
\caption{Surprise-adaptive agent}
\label{pseudocode}
\begin{algorithmic}[1]
\State Initialize network parameters $\phi$, replay buffer $\beta$, initial mean of bandit arms $\mu^{(0)}$, and initial optimization direction $\alpha^{(0)} \sim \text{Bern}(0.5)$
% \State $\alpha^{(0)} \sim \text{Bern}(0.5)$
\State Compute $H(p_{\theta_T}^{{\text{rand}}})$ by rolling out random trajectories \label{alg:rand}
\For{episode $m = 0,1 \dots,  \text{M}$}
% \State Initialize state marginal distribution parameters $\theta_0$
\State $s_o \sim p(s_0)$, \text{reset} $\theta_0$, $\bar{s_0} = (s_0, \theta_0, 0, \alpha^{(m)})$  \Comment construct initial augmented state
\State Set $r(s_{t}, a_t) = (-1)^{\alpha^{(m)}}-\log p_{\theta_t}(s_t)$ \Comment set reward function
        \For{$t = 0, \dots,  T$}
        \State Collect experience and update policy $\phi \gets RL(\phi, \beta)$
        \Comment See \citet{berseth2019smirl} \label{alg:rl}  
    \EndFor
    \State $\mu_i^{(m+1)} \gets \mu_i^{(m)} + \frac{1}{N(i)}(f_m - \mu_i^{(m)} ) $ if $\alpha^{(m)} = i$ else $\mu_i^{(m)}$ \label{alg:feedback}
    \State $\alpha^{(m+1)} \gets \text{UCB}(\mu^{(m+1)})$ \Comment Choose new $\alpha^{(m+1)}$ based on UCB algorithm \cite{lai1985asymptotically} \label{alg:select_bandit}
    \EndFor
\end{algorithmic}
\label{alg:surprise-adapt}
\end{algorithm}
% \vspace{-0.5cm}

\section{Experiments and analysis}
\label{sec:results}
% o evaluate \methodName, we (1) review the method's ability to control the entropy in the environment by showing \methodName will always select the best of s-max vs s-min. (2) Explore how this objective enables an agent to learn useful skills across a large collection of environments. (3) We review the components of S-Adapt to revise how to model the entropy best to result in a skilled agent ~\todo{this is our ablation analysis}.
%%GB.03.05.24: We can discuss the experiment to show the effect of different densities of butterflies/stochasticity.
% Experiments are conducted to validate the surprise-adaptive agent. First, 
%%GB.03.04.24: The above paragraph needs to be split into claims and how those are evaluated with some performance and a different paragraph on environments and prior methods.
To validate the usefulness and effectiveness of our method, we must demonstrate (1) Deficiencies in the single objective agents under particular entropy conditions and how these deficiencies arise from a lack of controllable entropy (2) Ability of the surprise-adaptive agent to select an objective based on the controllable entropy and to mimic the behavior of the single-objective agents and, finally (3) Correlation between entropy control and the emergence of intelligent behaviors.

 With these goals in mind, we select several environments for evaluation; First, a set of didactic environments that are designed specifically to create low- and high-entropy conditions to demonstrate both the success and failure modes of single-objective entropy control. Second, a set of RL benchmark environments that are not selected with any particular entropy conditions in mind, and hence are demonstrative of how our algorithm could perform on arbitrary environments with unknown entropy conditions. 
 %%03.07.24: These two paragraph should be connected. They seem to have the same topic.
 
 For the high-entropy environments, we select the \textit{Tetris} environment used in \citet{berseth2019smirl} and construct the new \textit{Butterflies} environment, shown in \Cref{fig:didactic_envs}. In \textit{Tetris} the agent must survive as long as possible by clearing rows of blocks before they reach the top of the frame. In \textit{Butterflies}, the agent (red) must find and catch butterflies (blue) that are moving randomly in the map, within a fixed-length episode. For the low-entropy environment, we construct a static maze environment (\textit{Maze}), in which the agent navigates around a map with a single goal state, for a fixed-length episode. For both \textit{Butterflies} and \textit{Maze}, we construct small (10x10) and large (32x32) versions of the environments. More details on these new environments are available in \Cref{sec:app-environments}. For the benchmark environments, we select the MinAtar \citep{young2019minatar} suite of tasks. This suite consists of simplified versions of five Atari games, which are designed to make the state space categorical and fully observable without frame-stacking. Finally, to experiment on image-based observations, we test on \textit{Freeway} from the original Atari games suite \citep{bellemare2013arcade}

% Our analysis puts the two dominant intrinsic reward objectives in contrast. To represent the space of curiosity-based methods we construct the \textbf{s-max} agent, that optimizes the local entropy of the agent's most stationary distribution~\citep{curiosity}. To represent the entropy minimizing regime \textbf{s-min}, a version of suprise-minimization is used~\cite {berseth2019entropy}. The choice of the above algorithms is also important as we construct \methodName to learn how to choose between the above objectives.
% We compare our method (\textbf{S-Adapt}) against an exclusively surprise-minimizing agent (\textbf{S-Min}), an exclusively surprise-maximizing agent (\textbf{S-Max}), an agent trained on the extrinsic reward (\textbf{Extrinsic}), and a random agent (\textbf{Random}). 
% The surprise maximizing agent \textbf{s-max} is an example of a \textit{curiosity}-based agent~\cite{}.
% We compare the performance of the various agents both in terms of entropy control and emergent behaviors. As a measure of entropy control, we consider the average surprise the agent experiences across the episode. The metric for emergent behavior that we consider here is the undiscounted episode return, as previous work has argued that entropy control can correlate with task rewards in some environments \citep{berseth2019smirl}. 

Our analysis contrasts our method with the two dominant entropy-based intrinsic reward paradigms. 
% To represent the space of curiosity-based methods ~\citep{pathak2017curiosity}, we construct the \textbf{S-Max} agent, that optimizes the entropy of the agent's state marginal distribution. To represent the entropy minimizing regime, we construct a surprise minimization agent \textbf{S-Min} ~\cite {berseth2019entropy}. 
Hence, we compare our method (\textbf{S-Adapt}) against an exclusively surprise-minimizing agent (\textbf{S-Min}) \citep{berseth2019smirl} and an exclusively surprise-maximizing agent (\textbf{S-Max}). Here, the surprise-maximizing agent represents the space of curiosity and maximum entropy methods \citep{pathak2017curiosity,hazan2019provably}, though we note that our method could be implemented on top of any desired curiosity-based method. Additionally, as baselines, we compare the entropy-based intrinsic agents against an agent trained on the extrinsic reward (\textbf{Extrinsic}), and a random agent (\textbf{Random}). 

We compare the performance of the various agents both in terms of entropy control and emergent behaviors. As a measure of entropy control, we consider the average surprise the agent experiences across the episode. The metric for emergent behavior that we consider here is the undiscounted episode return, as previous work has argued that entropy control can correlate with task rewards in some environments \citep{berseth2019smirl}.

\subsection{Failures of Single-Objective Entropy Control}
%%03.07.24: Good section title!
%
\label{sec:results_failure}
First, we demonstrate, qualitatively and quantitatively, the success and failure modes of single-objective entropy-based agents, using the didactic environments.

\begin{figure}[htb!]
\centering
\includegraphics[width=\textwidth]{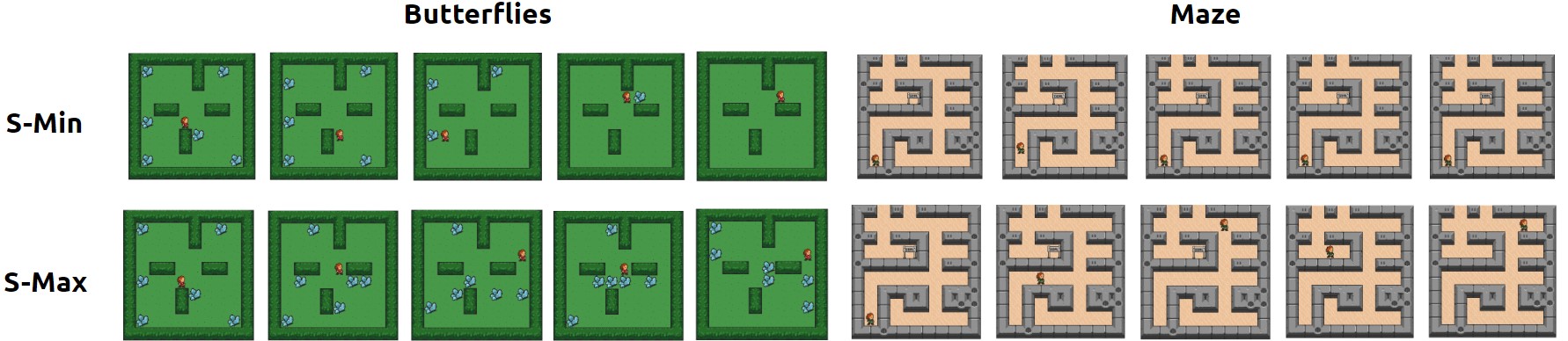}
\caption{The \textit{Butterflies} (left)  and  Maze environments (right). \textbf{S-Min} trains the agent to actively catch the butterflies in order to prevent diverse state configurations while at the same time preventing the agent to navigate around \textit{Maze}. \textbf{S-Max} trains the agent to avoid catching butterflies while navigating the Maze efficiently. These two didactic environments show that current intrinsic objectives fail to provide generally useful objectives for RL agents and cannot adapt. \label{fig:didactic_envs}}
\end{figure}

Qualitatively, we demonstrate the behaviors of \textbf{S-Min} and \textbf{S-Max} in the \textit{Maze} and \textit{Butterflies} environments in \Cref{fig:didactic_envs}. We note that the \textbf{S-Min} agent achieves an interesting behavior of catching butterflies in the \textit{Butterflies} environment, but learns a degenerate solution of standing in place in the \textit{Maze} environment. On the other hand, the \textbf{S-Max} agent learns to navigate the \textit{Maze} and reach the goal but fails to catch any butterflies in the \textit{Butterflies} environment.

\begin{figure}[htb!]
\begin{subfigure}{.49\textwidth}
  \centering
  \includegraphics[width=.98\linewidth]{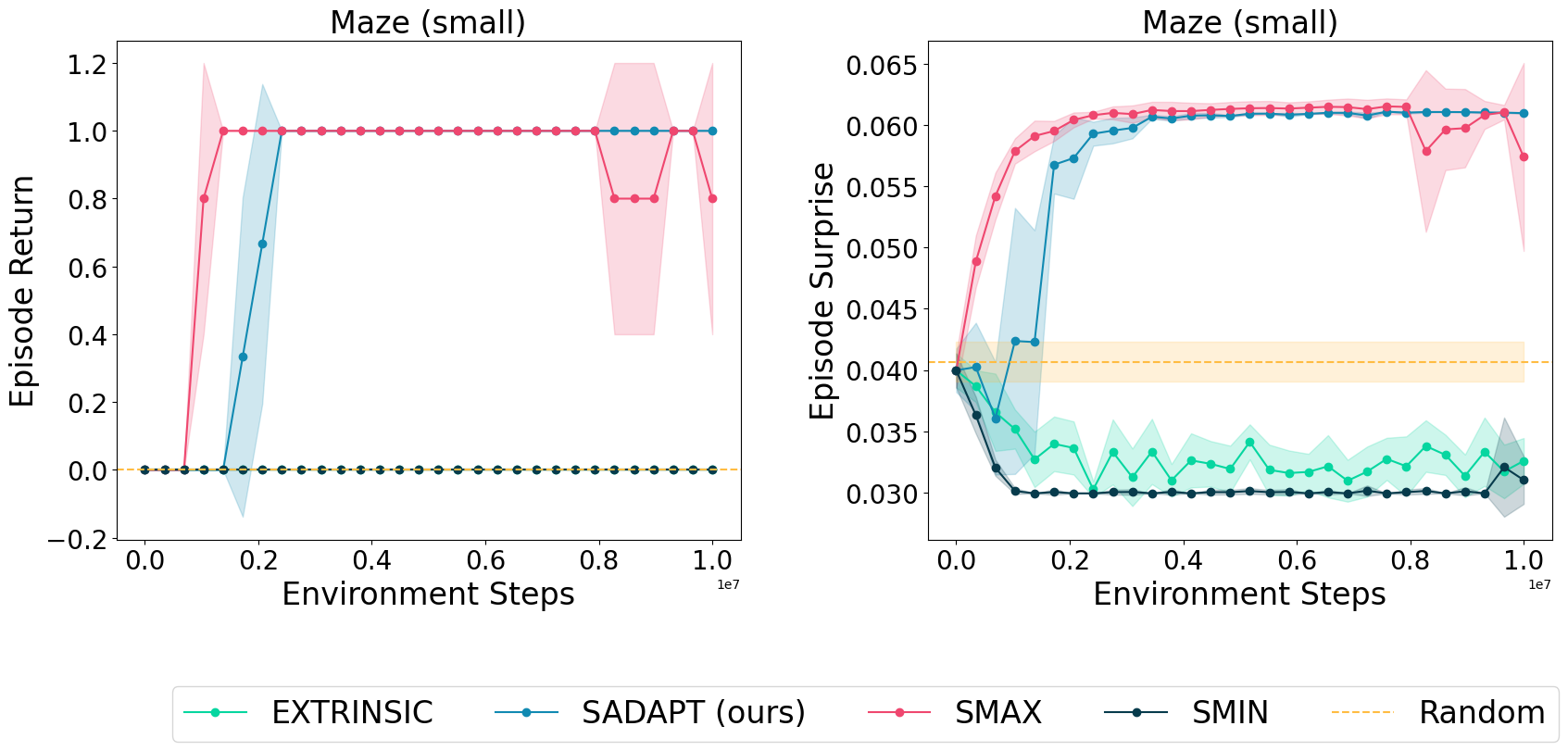}
  \caption{Small Maze (10x10)}
  \label{fig:maze_small}
\end{subfigure}%
\begin{subfigure}{.49\textwidth}
  \centering
  \includegraphics[width=.98\linewidth]{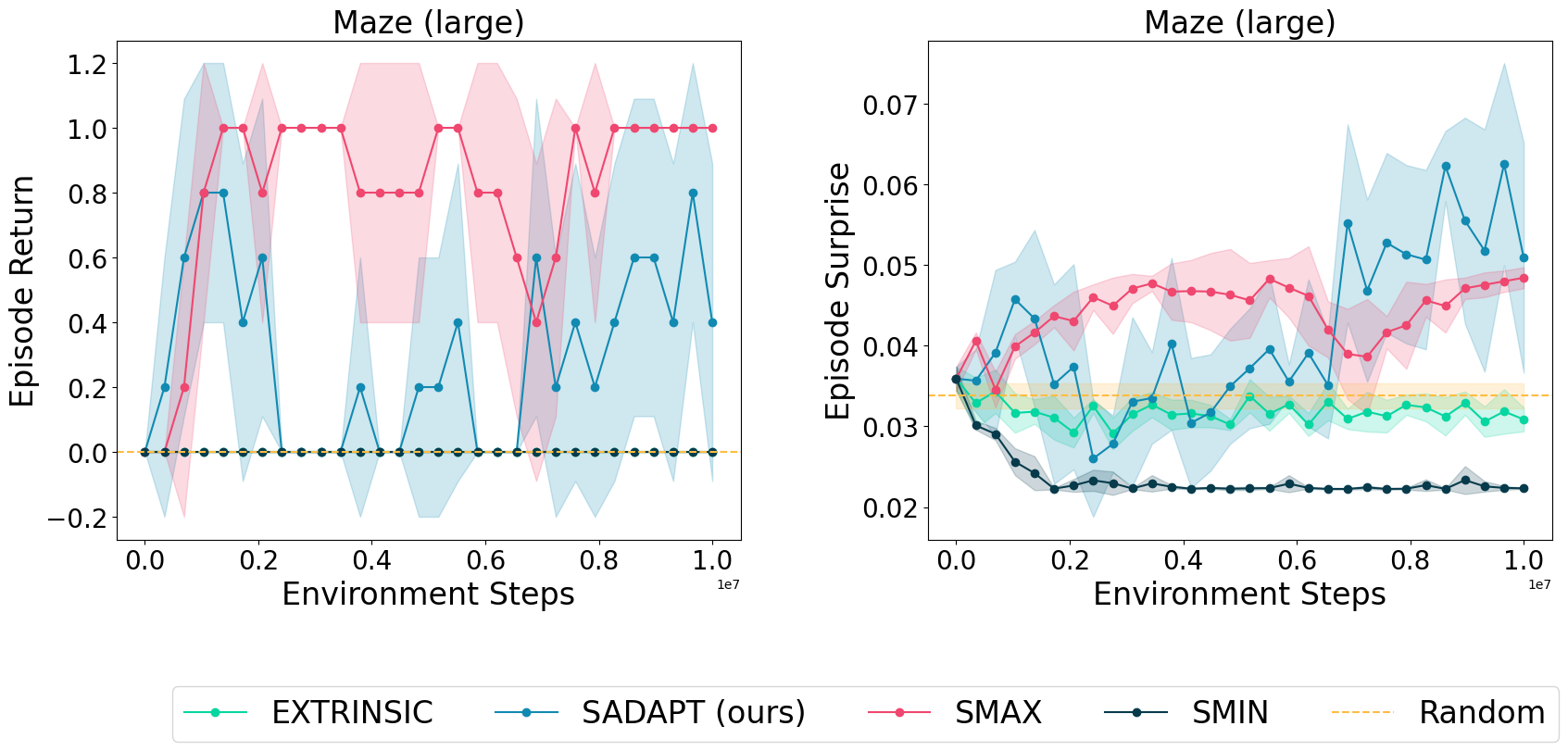}
  \caption{Large Maze (32x32)}
  \label{fig:maze_large}
\end{subfigure}
  \caption{Average episode return (left) and surprise (right) versus environment interactions (average over 5 seeds, with one shaded standard deviation) in the \textit{Maze} environment. \textbf{S-Max} and \textbf{S-Adapt} are the only objectives that allow the RL agents to consistently find the goal in the maze. These also cause the largest change in surprise when compared to the random agent.}
  \label{fig:maze}
\end{figure}

\begin{figure}[htb!]
\centering
\begin{subfigure}{.49\textwidth}
  \centering
  \includegraphics[width=.98\linewidth]{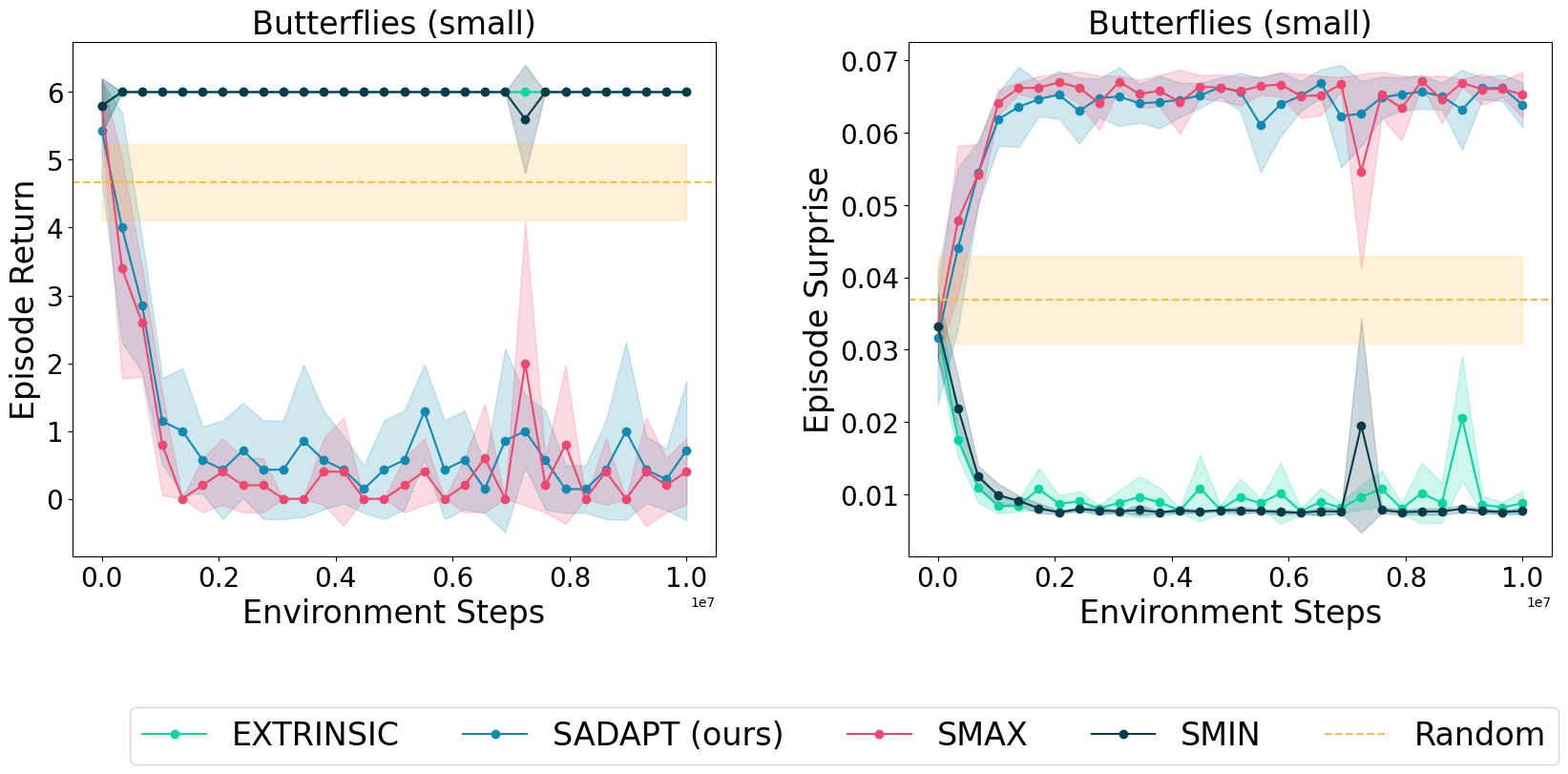}
  \caption{Small Butterflies (10x10)}
  \label{fig:butterflies_small}
\end{subfigure}%
\begin{subfigure}{.49\textwidth}
  \centering
  \includegraphics[width=.98\linewidth]{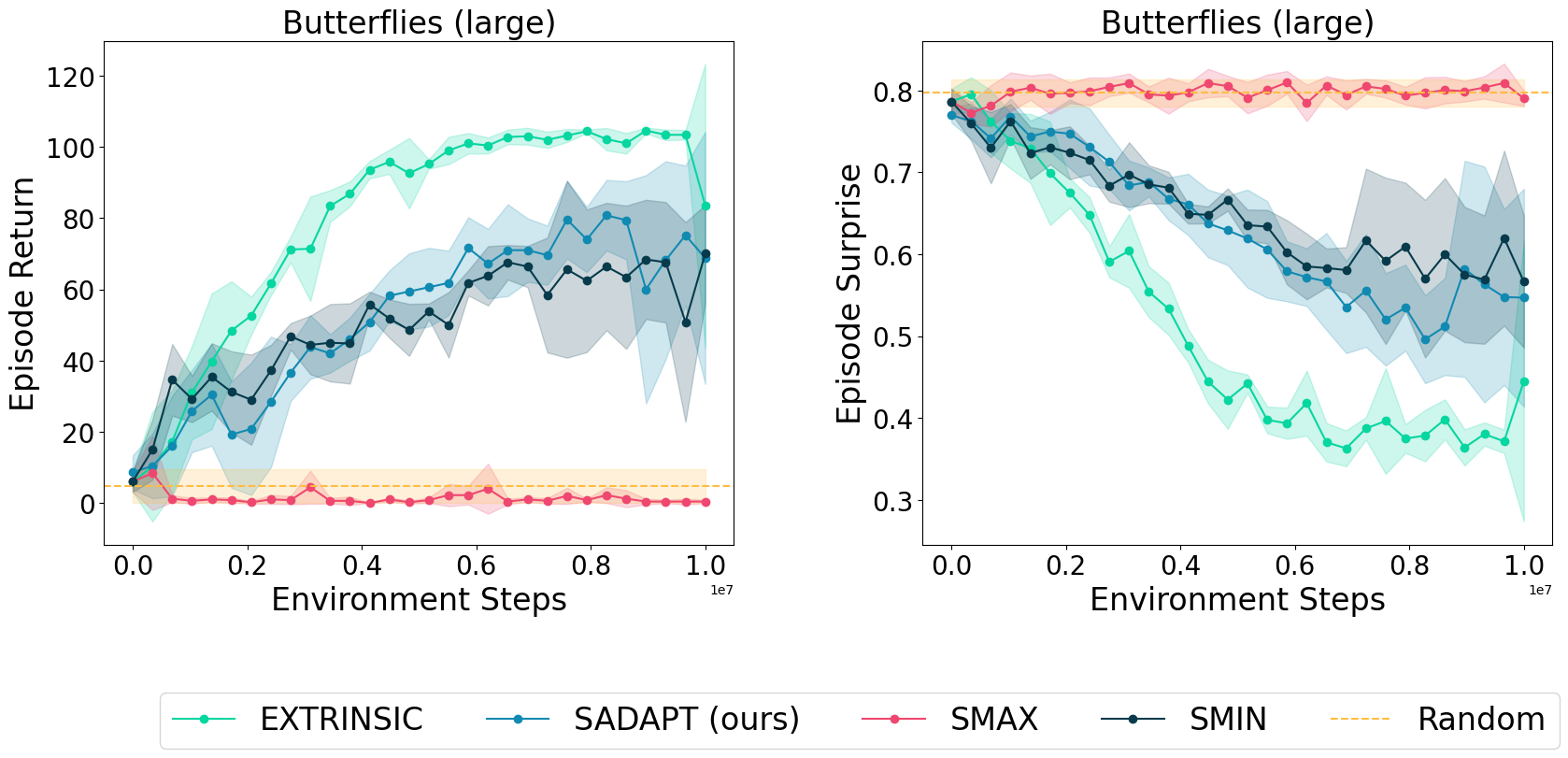}
  \caption{Large Butterflies (32x32)}
  \label{fig:butterflies_large}
\end{subfigure}
  \caption{Average episode return (left) and surprise (right) versus environment interactions (average over 5 seeds, with one shaded standard deviation) in the \textit{Butterflies} environment. \textbf{S-Min}, \textbf{Extrinsic} and even the \textbf{Random} agent catch most of the butterflies in the small grid. Because of the small size of the grid, surprise-minimization and surprise-maximization are equally effective in entropy control, and hence the  \textbf{S-Adapt} agent converges to \textbf{S-Max}. In the larger grid, however, the \textbf{Random} agent can't catch many butterflies and hence has a high-entropy state distribution. Again, the \textbf{S-Max} agent learns to also avoid catching butterflies and the \textbf{S-Min} agent learns to catch butterflies. However, catching butterflies results in a significant change in the state-marginal entropy in this larger grid. The \textbf{S-Adapt} agent identifies this and converges to \textbf{S-Min}, resulting in agents that catch more than half of the butterflies without access to the extrinsic reward.}
  \label{fig:butterflies}
\end{figure}

\begin{wrapfigure}{r}{0.5\textwidth}
\vspace{-1em}
  \centering
  \includegraphics[width=0.5\textwidth]
  {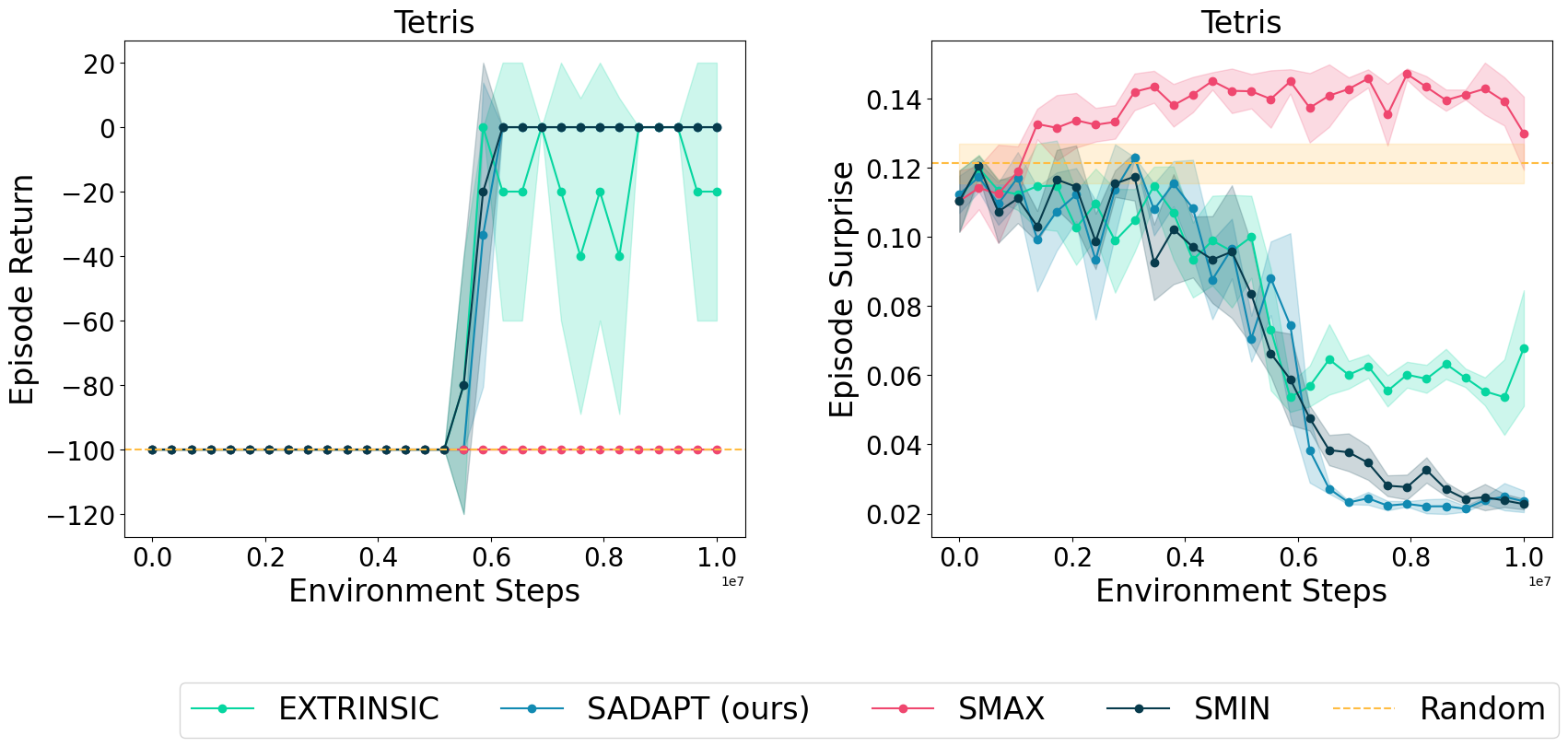}
  \caption{Average episode return (left) and surprise (right) versus environment interactions (average over 5 seeds, with one shaded standard deviation) in \textit{Tetris}. \textbf{S-Min}, \textbf{S-Adapt}, and \textbf{Extrinsic} agents solve the game (i.e. consistently survive for more than 200 steps). Interestingly, the surprise-minimizing objective, which \textbf{S-Adapt} converges to, turns out to be a better learning signal than the row-clearing extrinsic reward in \textit{Tetris}.}
  \label{fig:tetris}
\end{wrapfigure}

Quantitatively, we evaluate the average surprise and average extrinsic returns for the agents across training in all environments (\Cref{fig:maze,fig:butterflies,fig:tetris}). Notably, the \textbf{S-Min} agent achieves the lowest or near-lowest entropy in all environments, while the \textbf{S-Max} agent achieves the highest or near-highest entropy in all environments, as expected.

However, we highlight that the qualitatively interesting direction for entropy control is correlated not with a single objective, but with the scale of the absolute difference in the final entropy achieved by the agent versus that of the \textbf{Random} agent. For example, in the \textit{Maze} environment, the \textbf{S-Max} agent drives a significant increase in entropy over the \textbf{Random} agent, while the \textbf{S-Min} agent achieves a relatively small decrease (\Cref{fig:maze}). Similarly, in the \textit{Butterflies} environment, the opposite holds in the large map (\Cref{fig:maze_large}). Interestingly, in the small map, the \textbf{S-Min} and \textbf{S-Max} agents achieve roughly the same absolute change in entropy (\Cref{fig:maze_small}). This is because in the smaller map, avoiding butterflies is equally challenging compared to catching butterflies, while in the larger map, the butterflies are easily avoided.

\subsection{ Adaptive Entropy Control}
\label{sec:entropy-control}

Capitalizing on the success modes of the single-objective agents, the proposed \textbf{S-Adapt} agent can adapt to the entropy landscape to achieve entropy control across all didactic environments (\Cref{fig:maze,fig:butterflies,fig:tetris}). In \textit{Maze}, the \textbf{S-Adapt} agent converges to a surprise-maximizing strategy similar to \textbf{S-Max}, as demonstrated by the high entropy achieved by the end of training (\Cref{fig:maze}). On the other hand, in \textit{Tetris}, the \textbf{S-Adapt} agent converges to a surprise-minimizing strategy, achieving low entropy on par with the \textbf{S-Min} agent by the end of training (\Cref{fig:tetris}). In the \textit{Butterflies} environment, an interesting dichotomy in the \textbf{S-Adapt} agent's behavior arises. As noted in \Cref{sec:results_failure}, in the small grid, both the \textbf{S-Min} agent and \textbf{S-Max} agent induce roughly the same amount of change in the entropy versus the \textbf{Random} agent, using equally challenging strategies (\Cref{fig:butterflies_small}). Here, the \textbf{S-Adapt} agent converges to surprise-maximizing behavior. However, as the size of the grid is increased, and the density of butterflies decreases, the effect of minimizing entropy becomes much stronger versus the \textbf{Random} agent and the \textbf{S-Adapt} agent correctly converges to the surprise-minimizing strategy (\Cref{fig:maze_large}). More details on the effect of butterfly density on the behavior of the \textbf{S-Adapt} agent can be found in \Cref{sec:app-butterfly-density}

\begin{figure}[h!]
\centering
\begin{subfigure}{.49\textwidth}
  \centering
  \includegraphics[width=.98\linewidth]{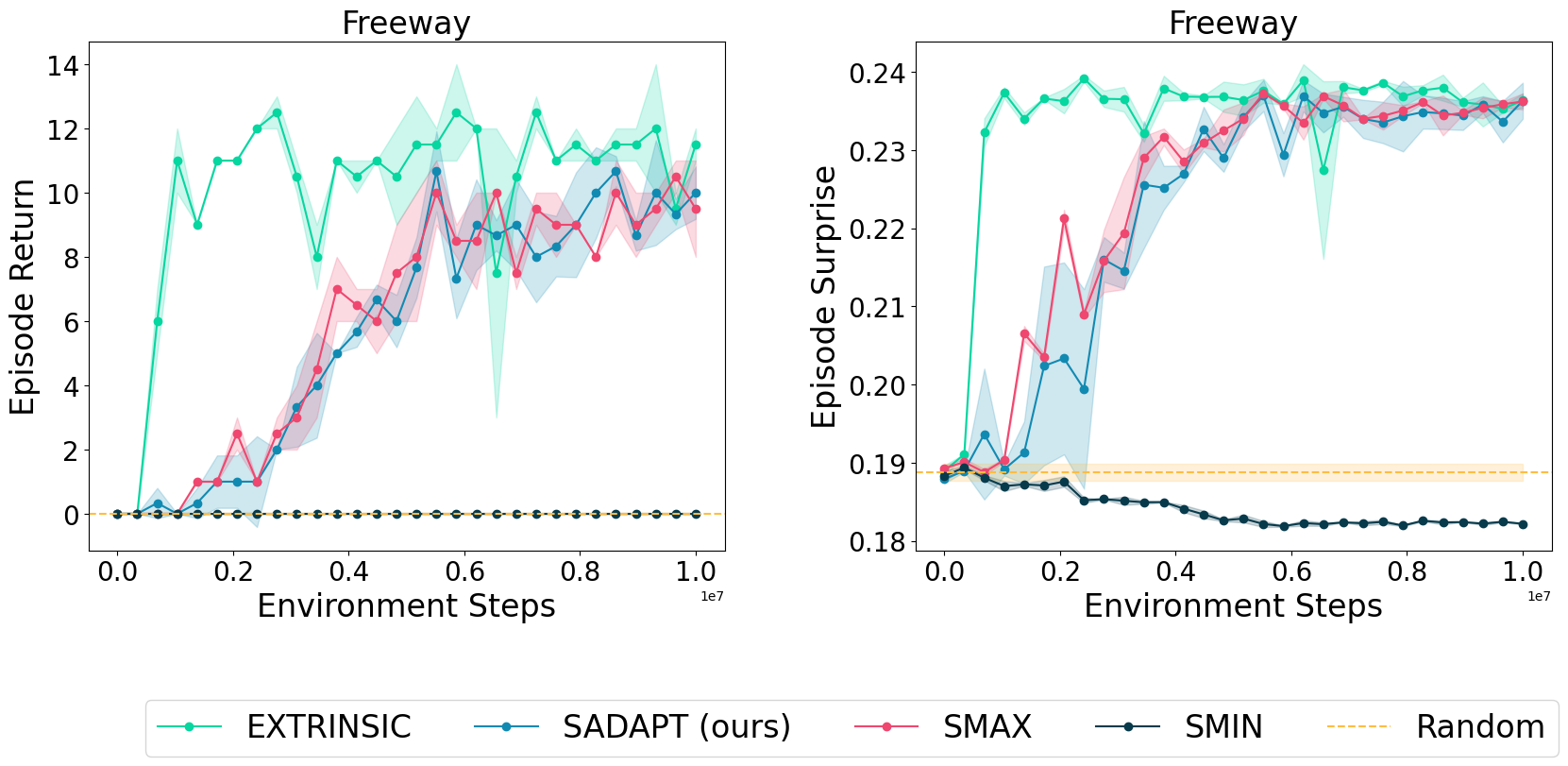}
  \caption{Freeway}
  \label{fig:freeway}
\end{subfigure}%
\begin{subfigure}{.49\textwidth}
  \centering
  \includegraphics[width=.98\linewidth]{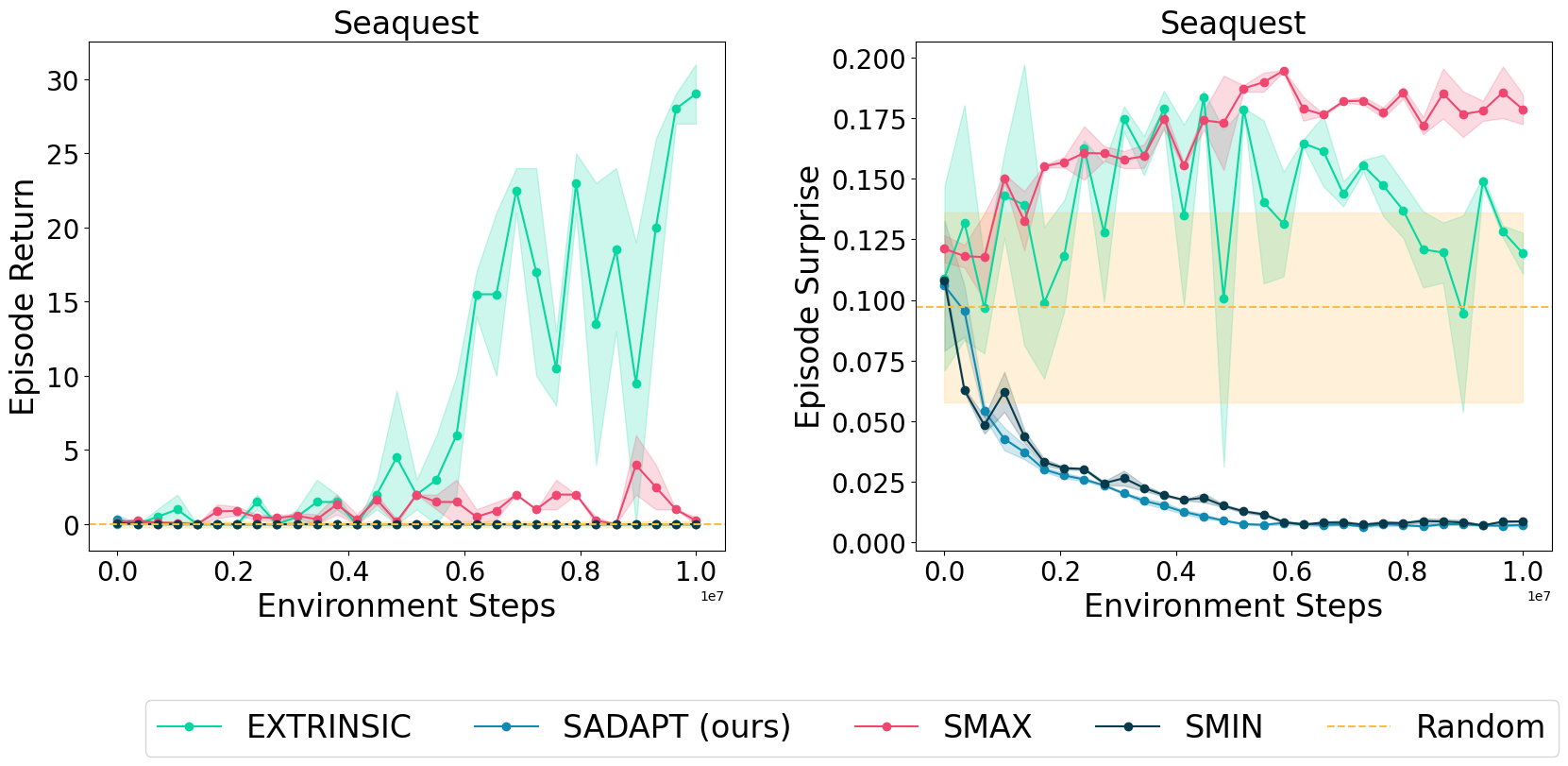}
  \caption{Seaquest}
  \label{fig:seaquest}
\end{subfigure}
\newline
\begin{subfigure}{.49\textwidth}
  \centering
  \includegraphics[width=.98\linewidth]{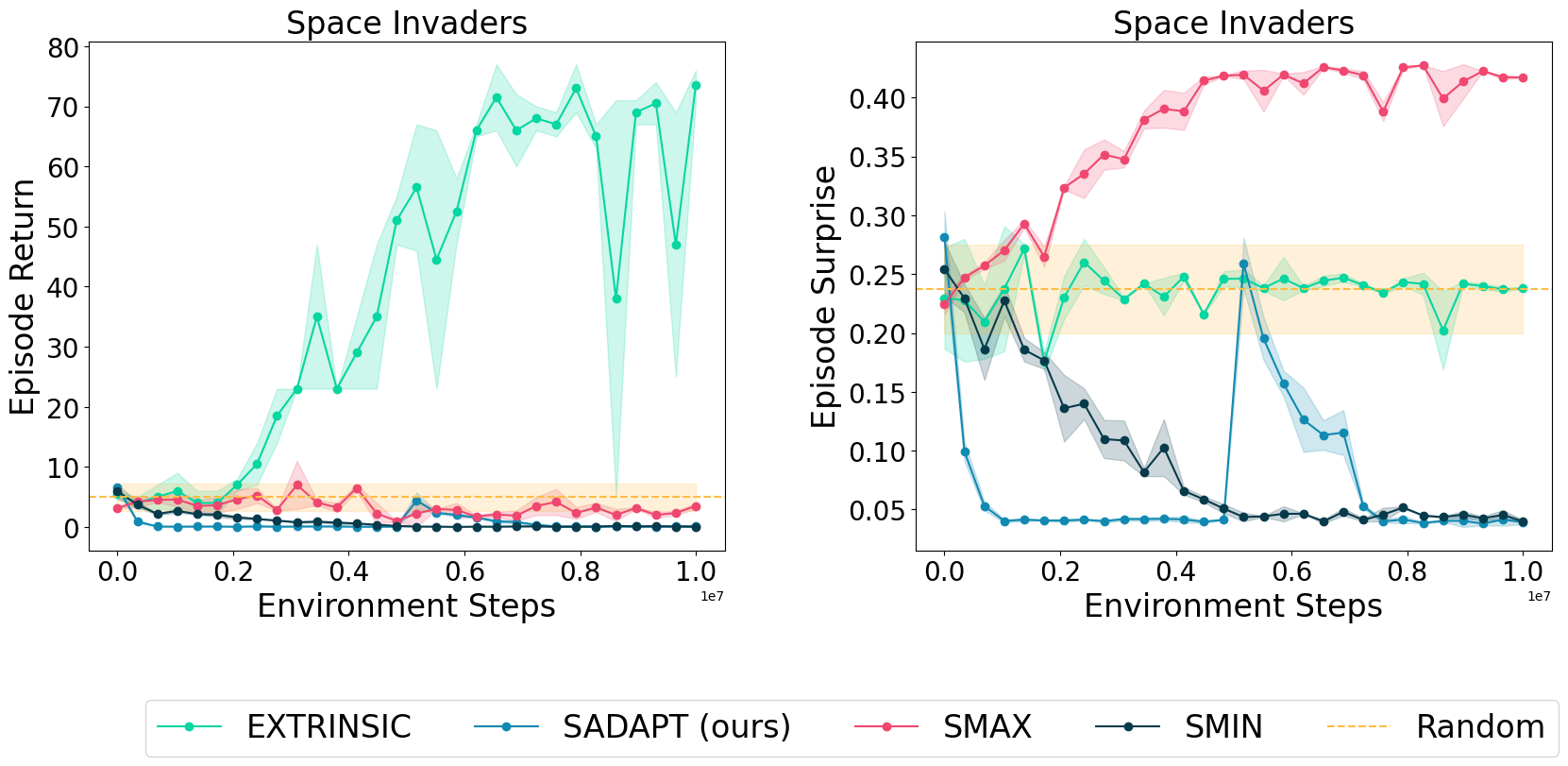}
  \caption{Space Invaders}
  \label{fig:space_invaders}
\end{subfigure}
\begin{subfigure}{.49\textwidth}
  \centering
  \includegraphics[width=.98\linewidth]{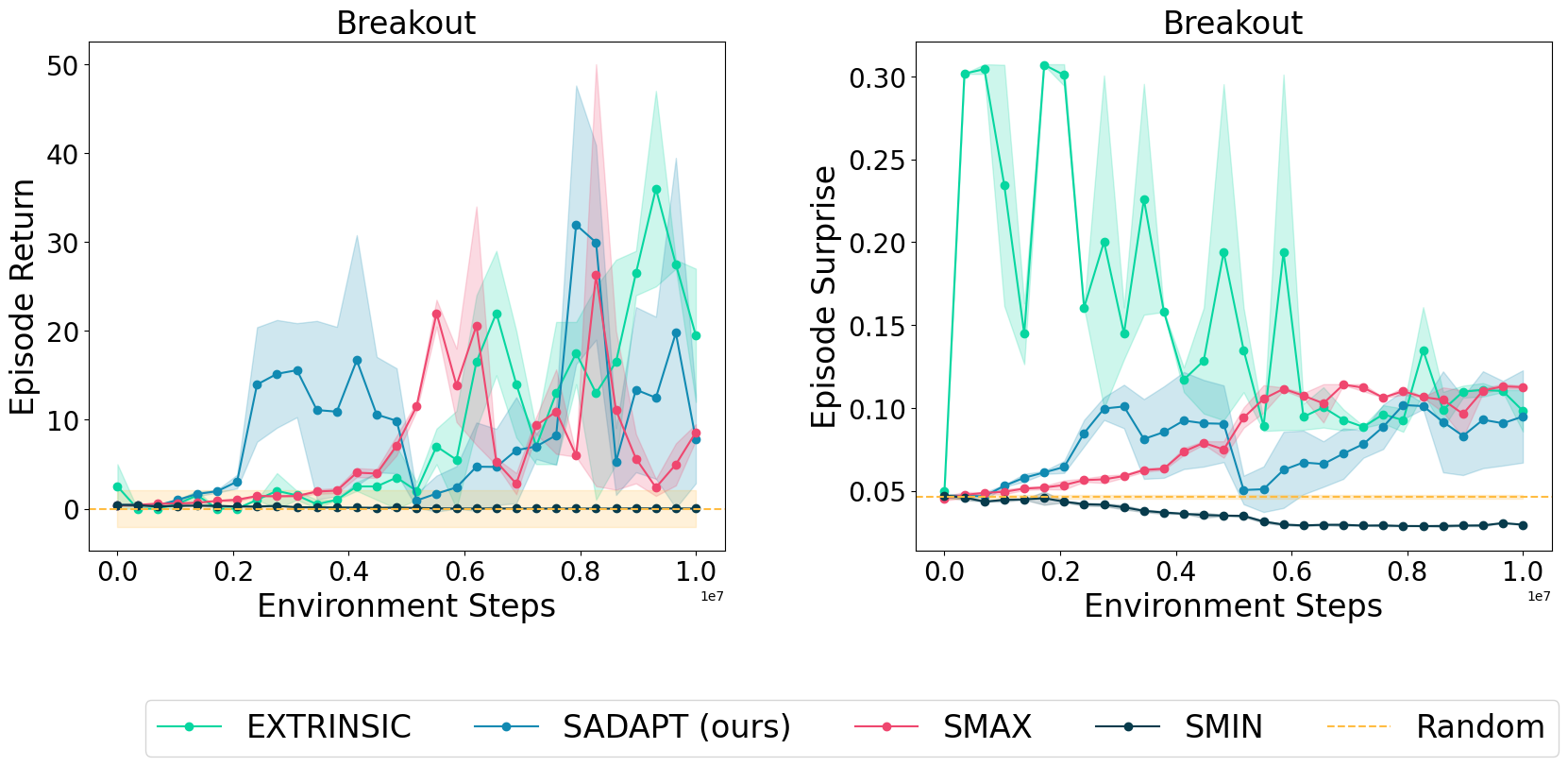}
  \caption{Breakout}
  \label{fig:breakout}
\end{subfigure}
\newline
\begin{subfigure}{.49\textwidth}
  \centering
  \includegraphics[width=.98\linewidth]{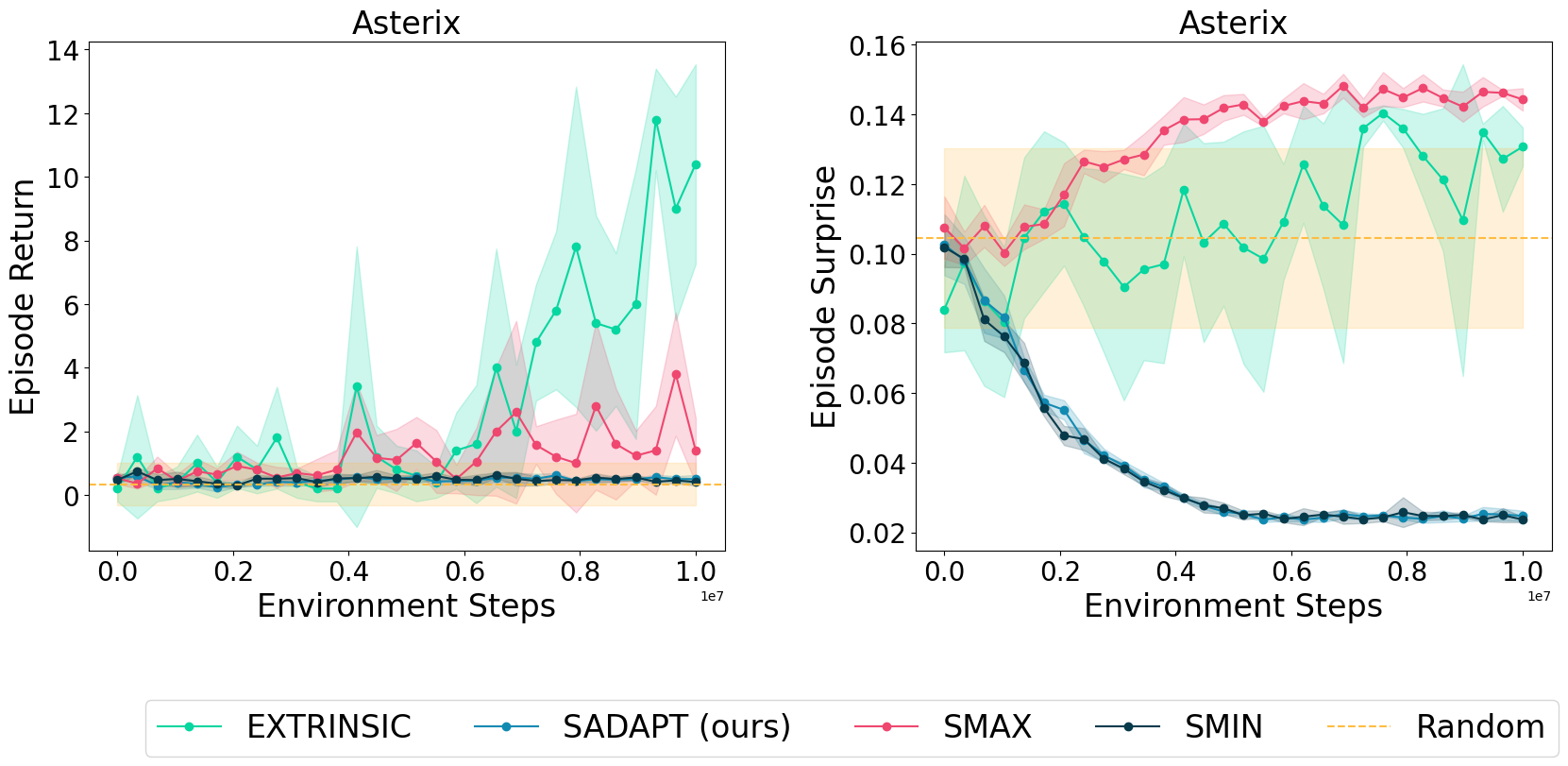}
  \caption{Asterix}
  \label{fig:asterix}
\end{subfigure}
  \caption{Average episode return (left) and surprise (right) versus environment interactions (average over 5 seeds, with one shaded standard deviation) in the MinAtar suite of environments. In all environments the \textbf{S-Adapt} agent is able to select the direction for entropy optimization which is most controllable, as demonstrated by the change in entropy from the beginning to the end of training. The \textbf{S-Adapt} agent indeed demonstrates emergent behaviors in certain environments, such as \textit{Freeway} where it achieves rewards on par with that of the \textbf{Extrinsic} agent. However, in certain environments, like \textit{Seaquest}, \textit{Space Invaders} and \textit{Asterix}, the extrinsic reward is not closely correlated with entropy control, with the \textbf{Random} agent and the \textbf{Extrinsic} agent achieving similar entropy.}
  \label{fig:minatar}
\end{figure}

Our results have shown that the \textbf{S-Adapt} agent can successfully recreate the performance of the \textbf{S-Min} and the \textbf{S-Max} agents in their respective didactic environments. Next, we investigate controlling entropy across the MinAtar benchmark, shown in \Cref{fig:minatar}. Notably, these environments were not constructed with any particular entropy regime in mind. Thus, these results are demonstrative of how the proposed algorithm could perform in an arbitrarily chosen environment. 

Here again, we see that the \textbf{S-Adapt} agent can reliably select the objective with the greatest controllable entropy. Though the difference between \textbf{S-Min} and \textbf{S-Max} agents in terms of divergence with the \textbf{Random} agent is not as strong in some environments, the \textbf{S-Adapt} agent consistently chooses the objective with the relatively larger change in entropy. This provides confirmation that our bandit algorithm can successfully select for controllable entropy in arbitrary environments.

%% GB: ALso, if we can get the sotry back for space invadres. We want some example where S-Adapt does better than other methods.

\begin{wrapfigure}{r}{0.5\textwidth}\centering
\begin{subfigure}{.49\textwidth}
  \centering
  \includegraphics[width=.98\linewidth]{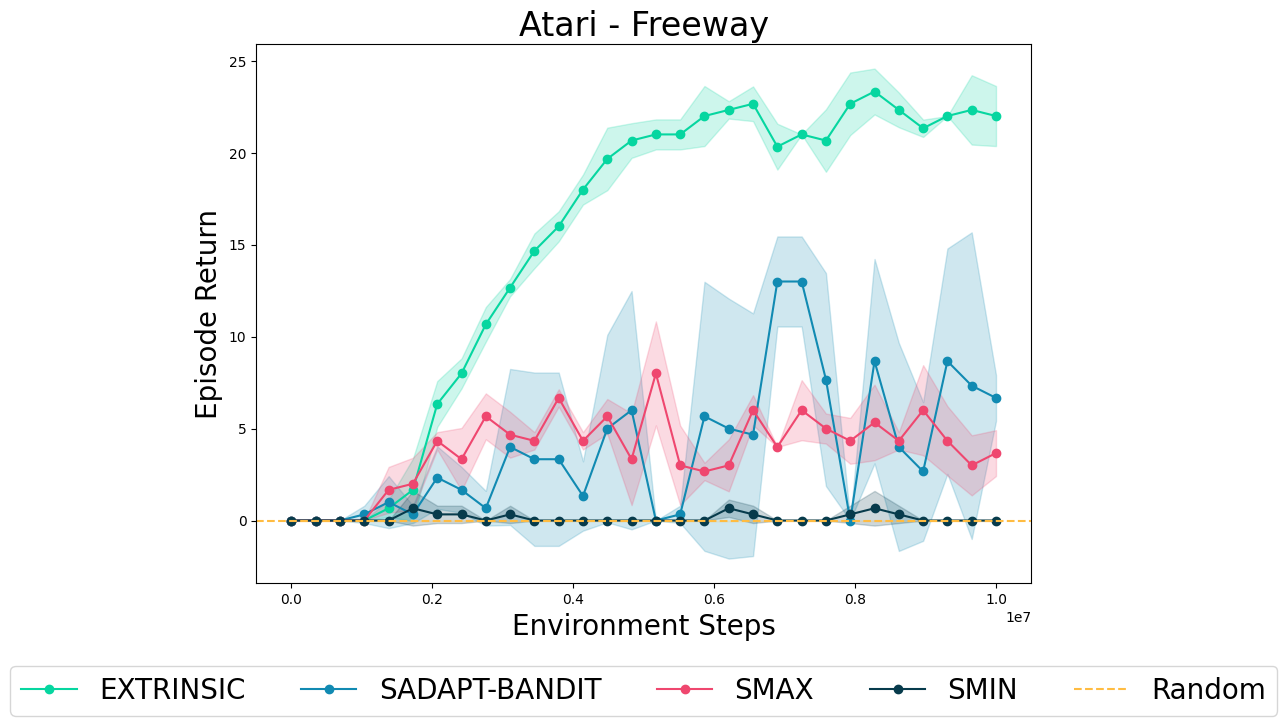}
  \label{fig:atari_freeway}
\end{subfigure}%
\caption{Average episode return versus environment interactions (average over 5 seeds, with one shaded standard deviation) in the Atari Freeway environment. The \textbf{S-Adapt} agent learns useful behaviours (making progress in the original task) from image-based observations. The \textbf{Extrinsic} agent achieves the highest returns as it exploits the task rewards, the \textbf{S-Max} agent achieves slightly lower returns than the \textbf{S-Adapt} agent, while the \textbf{S-Min} agent achieves zero returns.}
\label{fig:atari}
\end{wrapfigure}

\subsection{Emergent Behaviour}

Finally, for these objectives to be useful, it is important that they correlate with the emergence of interesting behaviors. Indeed, we note that the extrinsic rewards in the didactic environments generally correlate closely with one of two single-objective agents (\Cref{fig:maze,fig:butterflies,fig:tetris}). This suggests that these environments have good potential for entropy-based control to elicit emergent behaviors. Importantly, however, the extrinsic reward does not correlate well with strictly one of \textbf{S-Min} or \textbf{S-Max} in \textit{all} environments. In \textit{Maze}, \textbf{S-Max} achieves high rewards, while in \textit{Butterflies} and \textit{Tetris}, \textbf{S-Min} achieves high rewards. On the other hand, the \textbf{S-Adapt} agent achieves high task rewards, on par or better than the \textbf{Extrinsic} agent across \textit{all} didactic environments.

Additionally, in some MinAtar environments, the entropy-based agents exhibit emergent behavior similar to that of the \textbf{Extrinsic} agent. In the \textit{Freeway} environment (\Cref{fig:freeway}), the \textbf{S-Adapt} agent achieves competitive rewards with the \textbf{Extrinsic} agent. A similar result is observed in \textit{Breakout} (\Cref{fig:breakout}). However, other environments, like \textit{Space Invaders} and \textit{Seaquest} (\Cref{fig:space_invaders,fig:seaquest}) do not appear to be good candidates for intrinsic entropy control, since the \textbf{Extrinsic} and  \textbf{Random} agents achieve similar entropy.

Finally, we investigate the emergence of interesting behaviors in a more complex, image-based environment using \textit{Atari Freeway} (\Cref{fig:atari}) as a case study. Unlike the previous environments, observations in pixel space are non-binary and hence cannot be modeled using Bernoulli distributions. Instead, we model the state marginal using a Gaussian distribution (see \Cref{sec:est_state_marg} for more details). The results show that both the \textbf{S-Max} and \textbf{S-Adapt} agents achieve respectable results as compared to the extrinsic agent. Moreover, in this environment, the emergent behavior of the \textbf{S-Adapt} agent is qualitatively different from both \textbf{S-Max} and \textbf{S-Min} agents; The \textbf{S-Adapt} agent solves the game more frequently than the \textbf{S-Max} agent. This hints that mixing entropy maximization and minimization in one adaptive objective induces emergent behaviors that cannot be learned by exclusively optimizing for surprise minimization or maximization alone.

\section{Conclusion}
Our experiments demonstrate encouraging results for a surprise-adaptive agent. The \textbf{S-Adapt} agent can select the objective with the more controllable landscape across both didactic environments and benchmark environments. Moreover, the \textbf{S-Adapt} agent inherits the emergent behaviors of the single-objective agents, achieving high rewards across all didactic environments, which neither of the single-objective agents nor the extrinsic agent is able to achieve. Further work is needed to understand exactly under what conditions such emergent behaviors can manifest, and how to elicit them more reliably in arbitrary environments like MinAtar. Possible directions for improvement here could include better methods for estimating the state marginal distribution with more accuracy. Moreover, an interesting extension to this work would be to apply an adaptive agent in the continual learning setting, where adaptation can occur at any time, not only at episode end.

\subsubsection*{Acknowledgments}
\label{sec:ack}
We want to acknowledge funding support from NSERC, FRQNT, and CIFAR and compute support from the Digital Research Alliance of Canada, Mila IDT and NVidia.

%%%%%%%%%%%%%%%%%%%%%%%%%%%%%%%%%%%%%%%%%%%%%%%%%%%%%%%%%%%%%%%%
%% NOTE: THIS MARKS THE END OF THE "MAIN TEXT"
%%%%%%%%%%%%%%%%%%%%%%%%%%%%%%%%%%%%%%%%%%%%%%%%%%%%%%%%%%%%%%%%

%%%%%%%%%%%%%%%%%%%%%%%%%%%%%%%%%%%%%%%%%%%%%%%%%%%%%%%%%%%%%%%%
%% Bibliography
%%%%%%%%%%%%%%%%%%%%%%%%%%%%%%%%%%%%%%%%%%%%%%%%%%%%%%%%%%%%%%%%
\bibliography{references}
\bibliographystyle{rlc}

%%%%%%%%%%%%%%%%%%%%%%%%%%%%%%%%%%%%%%%%%%%%%%%%%%%%%%%%%%%%%%%%
%% Appendices
%%%%%%%%%%%%%%%%%%%%%%%%%%%%%%%%%%%%%%%%%%%%%%%%%%%%%%%%%%%%%%%%
\newpage
\appendix

\section{Environment and Training Details}
\label{appendix}
\subsection{Training Details}
All agents were trained using DQN \citep{mnih2015playing}. Reward values are normalized by subtracting the rolling mean and dividing by the standard deviation before fitting the Q network. For the \textbf{S-Adapt} agent, we use the original UCB algorithm with exploration coefficient 2 in the \textit{Maze} (large) and MinAtar environments, for all other environments we set the exploration coefficient to $\sqrt{2}$. we trained all agents using the implementation of DQN from CleanRL \citep{huang2022cleanrl}. We trained all agents with a learning rate of 0.0001 with Adam optimizer, a discount factor of 0.99, a batch size of 32, a replay buffer size of 1M, and for 10M environment interactions. We use epsilon-greedy for exploration with a linearly decaying epsilon from a value of 1 to 0.01, decaying over the first 10\% of timesteps in all environments except MinAtar and Atari which decays over the first 50\% of time steps. Model architecture details for each environment are provided in the next section.

\subsection{Environments}
\label{sec:app-environments}

\paragraph{Tetris}
We take the \textit{Tetris} environment directly from the implementation provided by the authors of \citep{berseth2019smirl}. In this environment, the agent receives 0 at all steps, except for a losing step which results in a -100 reward. The maximum episode length is 200. Environment observations and the sufficient statistic of the state marginal are flattened before being fed into two independent two-layer MLPs with hidden dimensions 120 and 84. The outputs of the MLPs are concatenated and passed through a linear layer that outputs the Q-value.

\begin{figure}[tbh]
\centering
\includegraphics[width=1\textwidth]{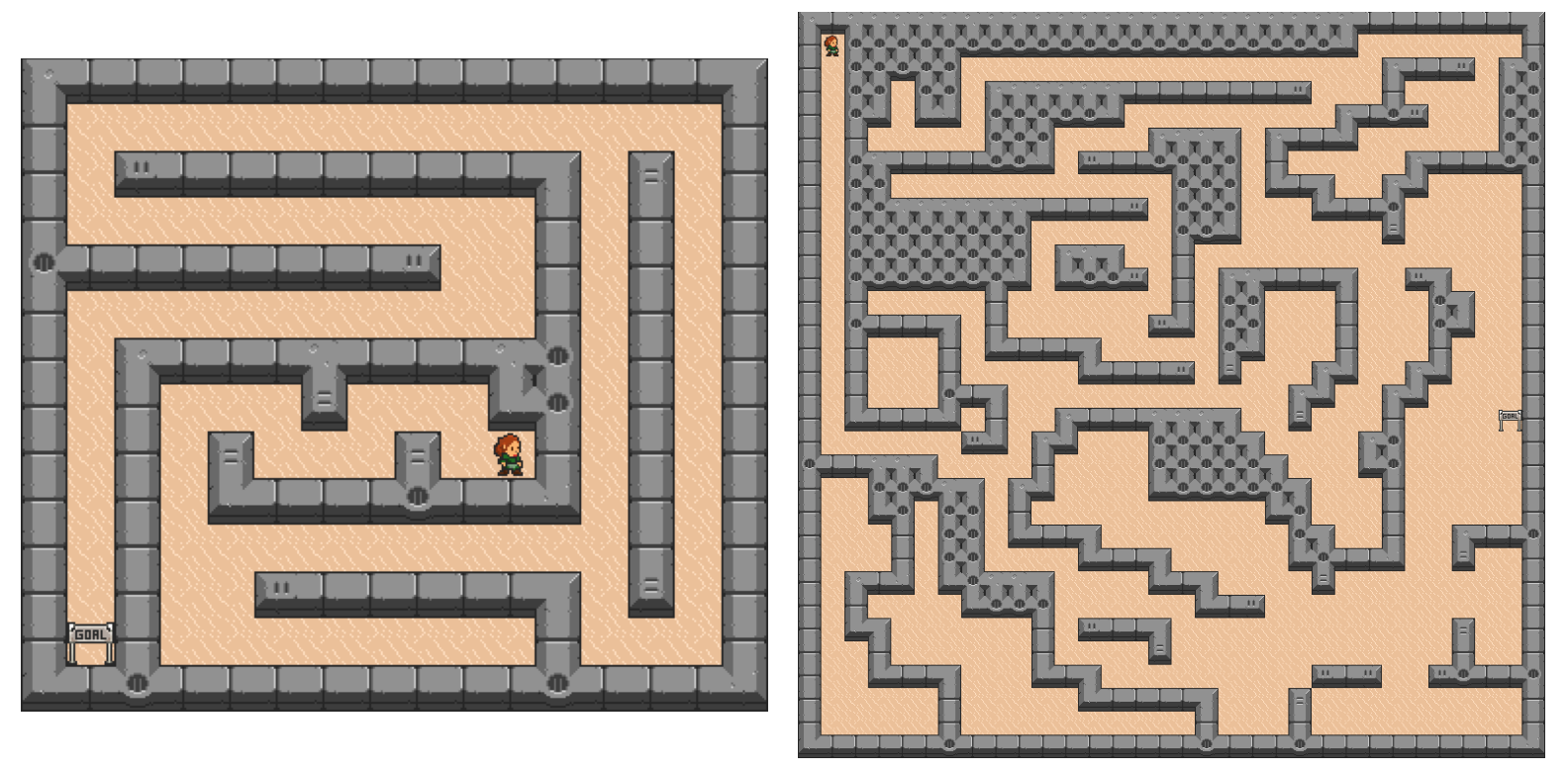}
\caption{Pixel-rendering of the small maze (left) and the large maze (right)}
\label{fig:maze_pic}
\end{figure}

\paragraph{Maze}
We constructed custom \textit{Maze} environments (small and large) using the Griddly platform \citep{bamford2021griddly}. A pixel-rendering of the small and large mazes used in our experiments can be found in \autoref{fig:maze_pic}. The task reward in both environments is +1 when the agent reaches the goal and 0 otherwise.

The size of the small maze is 10x10 and the episode length is 100. Environment observations and the sufficient statistic of the state marginal are passed through two independent CNNs with a single convolutional layer. The outputs of the CNNs are concatenated and passed through a single-layer MLP with hidden dimension 512 that outputs the Q-value.

The size of the large maze is 32x32 and the episode length is 250. Environment observations and the sufficient statistic of the state marginal are passed through two independent CNNs with three convolutional layers with kernel size of (3,3), a stride value of 2 and a padding value of 1. The outputs of the CNNs are concatenated and passed through a single-layer MLP with hidden dimension 512 that outputs the Q-value.

\paragraph{Butterflies}
We constructed the custom \textit{Butterflies} environment (small and large) using the Griddly platform \citep{bamford2021griddly}. The task reward in both environments is +1 when the agent catches a butterfly and 0 otherwise. 

The size of the small map is 10x10 and the episode length is 100, while the size of the large map is 32x32 and the episode length is 500.  We use the same architecture as the \textit{Maze} environment for estimating the Q-value.

\paragraph{MinAtar}
In MinAtar environments, we use the same architecture as the \textit{Butterflies} environments and we set the episode length to 500. 

\paragraph{Atari}
In Atari \textit{Freeway} environment, we use the same architecture and pre-processing as in \cite{mnih2015playing}. We use the same multiple CNN architecture as the \textit{Maze} environment for estimating the Q-values from the augmented state with sufficient statistics.

\newpage

\subsection{Estimation of State Marginal Distribution}
\label{sec:est_state_marg}
In all binary environments (\textit{Tetris, Maze, Butterflies, MinAtar}), the observed state $s_t$ is a binary entity map of size $H \times W \times C$, where $H$ is the height of the map, $W$ is the width of the map and $C$ is the number of channels, with each channel representing a single object type in the environment. A value of one is set in the $(h,w)$ position of channel $c$ (denoted $s_{t}^{h,w,c}$) if an object of type $c$ currently occupies the $(h,w)$ position in the map, and zero otherwise. The state marginal distribution is estimated as $H \times W \times C$ independent Bernoulli distributions, with probability $p_t^{h,w,c} = \frac{\sum_{t'=0}^t s^{h,w,c}_{t'}}{t}$, which constitutes a sufficient statistic for the Bernoulli distribution. Hence, the sufficient statistic of the entire state marginal distribution is given by $\theta_t = \{p_t^{h,w,c}: h\in H, w\in W, c \in C\}$ and is the same shape as the observations $s_t$. 

\begin{figure}[h!]
\centering
\begin{subfigure}{.33\textwidth}
  \centering
  \includegraphics[width=.98\linewidth]{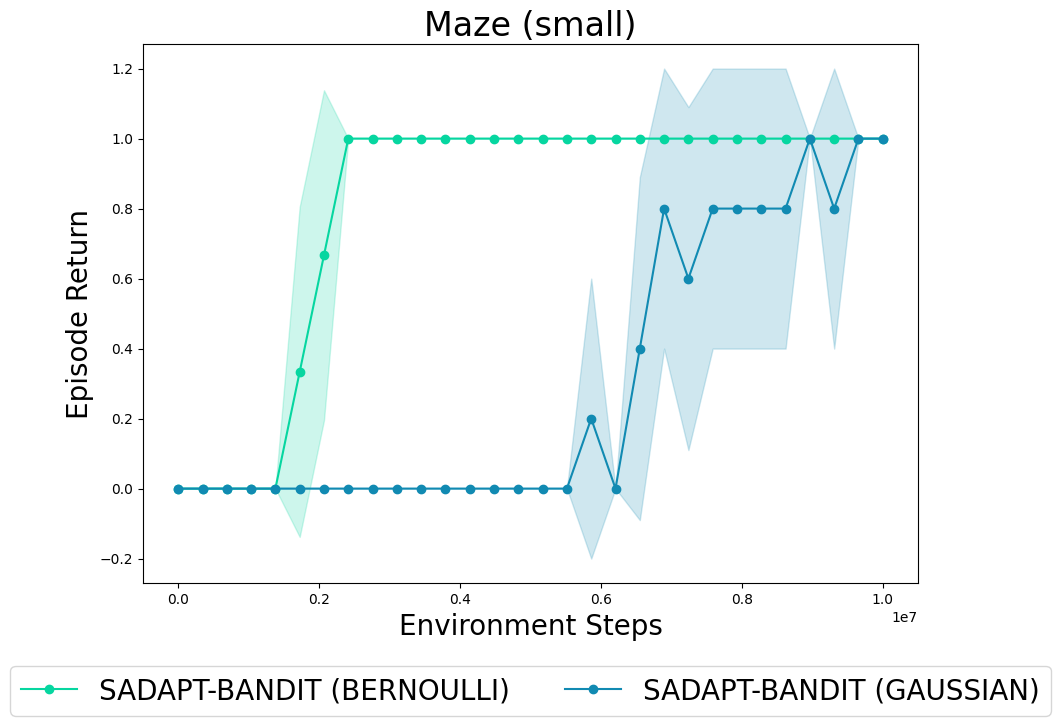}
  \caption{Maze (small)}
  \label{fig:maze_small_gauss}
\end{subfigure}%
\begin{subfigure}{.33\textwidth}
  \centering
  \includegraphics[width=.98\linewidth]{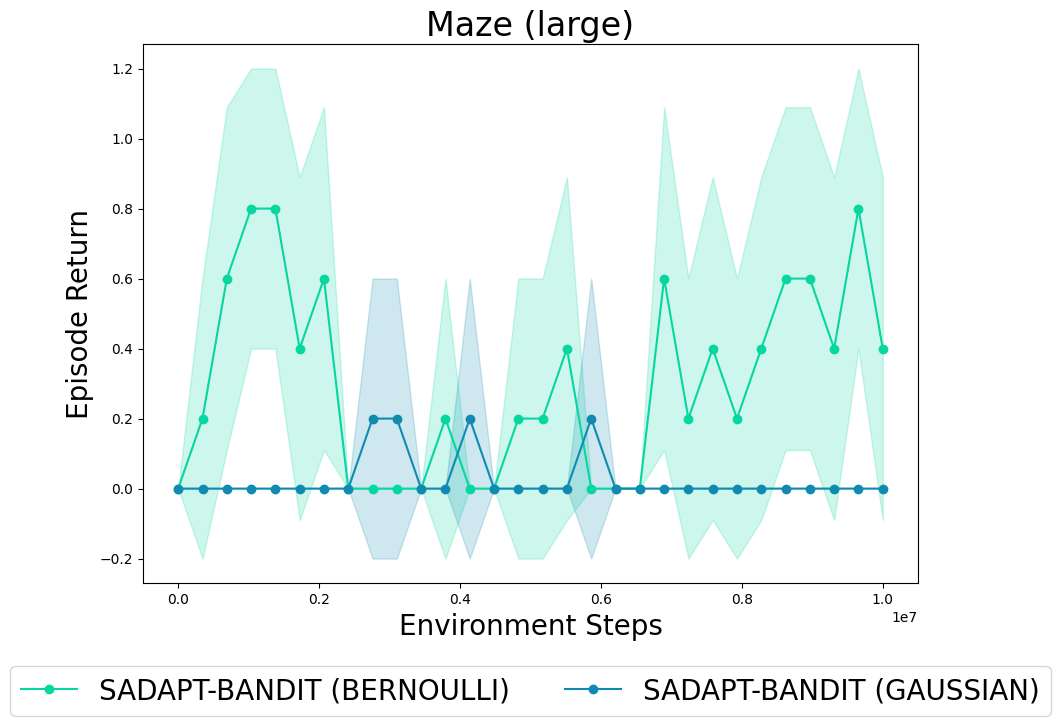}
  \caption{Maze (large)}
  \label{fig:maze_large_gauss}
\end{subfigure}
\begin{subfigure}{.33\textwidth}
  \centering
  \includegraphics[width=.98\linewidth]{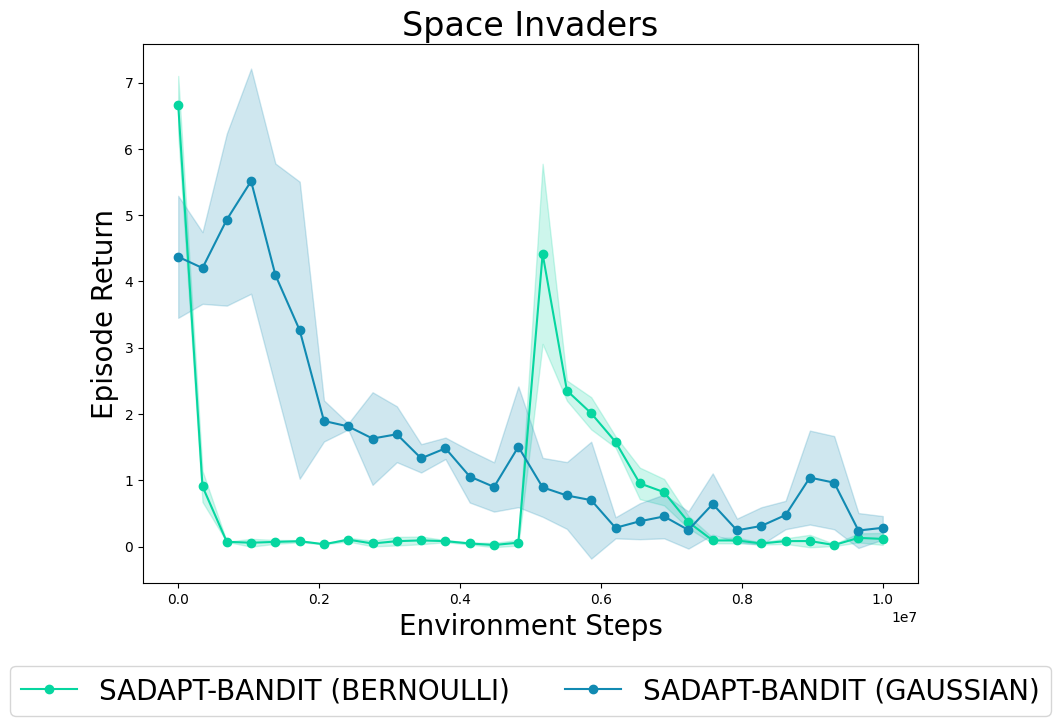}
  \caption{Space Invaders}
  \label{fig:space_invaders_gauss}
\end{subfigure}
\newline
\begin{subfigure}{.33\textwidth}
  \centering
  \includegraphics[width=.98\linewidth]{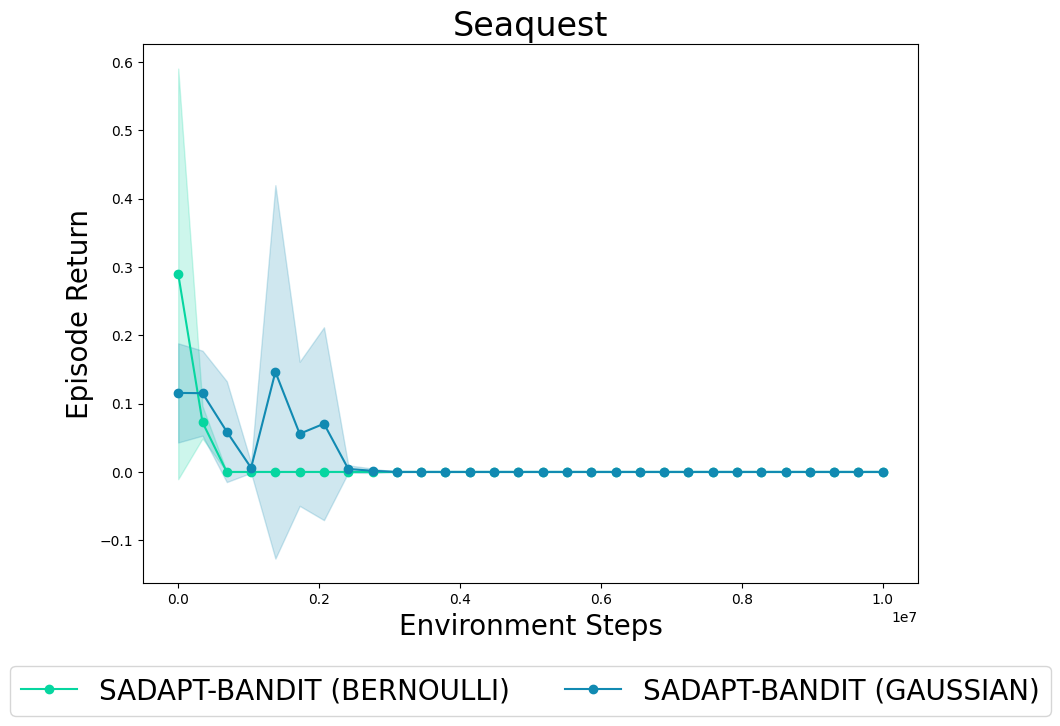}
  \caption{Seaquest}
  \label{fig:seaquest_gauss}
\end{subfigure}
\begin{subfigure}{.32\textwidth}
  \centering
  \includegraphics[width=.98\linewidth]{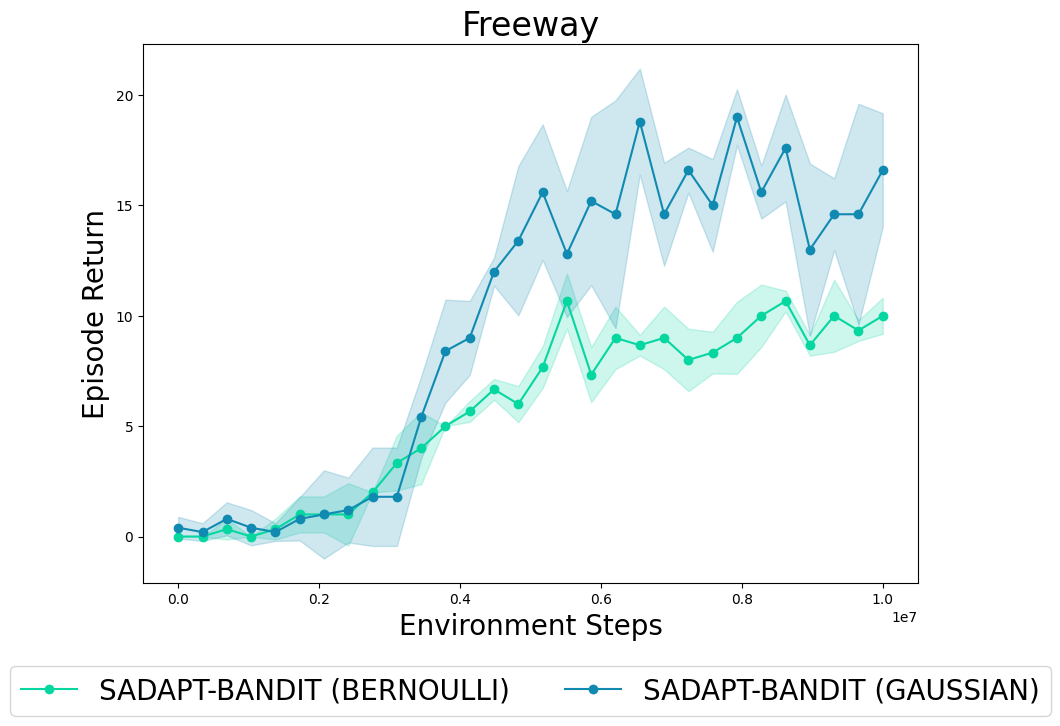}
  \caption{Freeway}
  \label{fig:freeway_gauss}
\end{subfigure}
\begin{subfigure}{.32\textwidth}
  \centering
  \includegraphics[width=.98\linewidth]{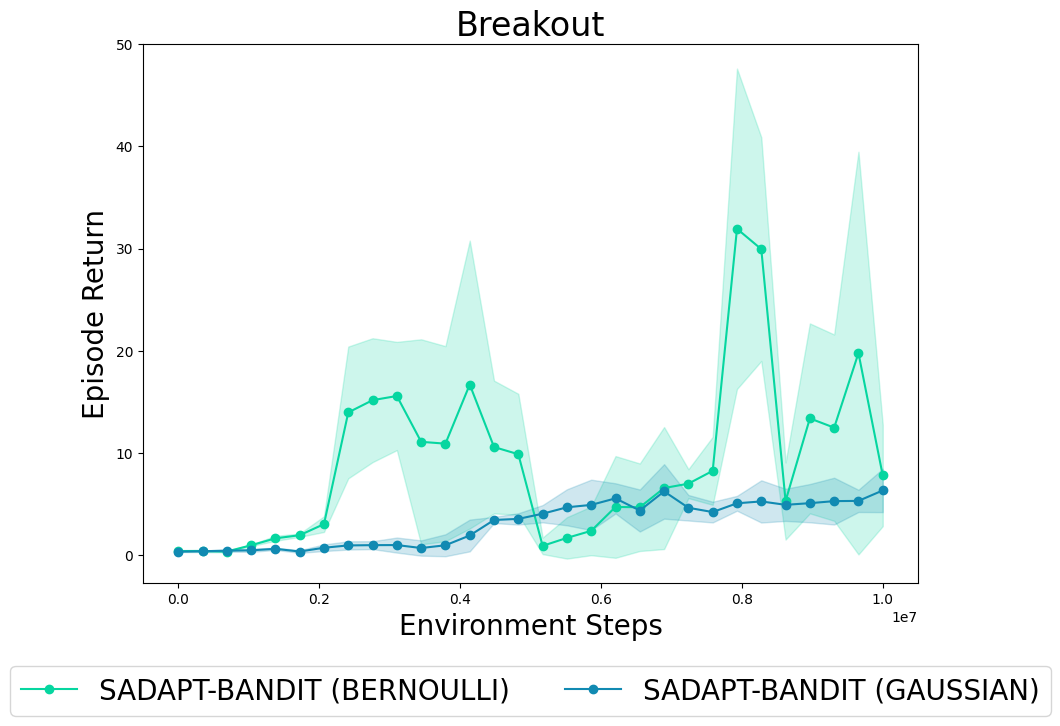}
  \caption{Breakout}
  \label{fig:breakout_gauss}
\end{subfigure}
  \caption{Average episode return of the \textbf{S-Adapt} (average over 5 seeds, with one shaded standard deviation), using Gaussian and Bernoulli distributions for estimating the state marginal distribution.}
  \label{fig:ablation}
\end{figure}

The choice of the Bernoulli distribution is justified by the binary nature of the data. However, we perform an ablation using a Gaussian distribution as an alternative to confirm the validity of this choice (\Cref{fig:ablation}).

In the image-based environment (Atari \textit{Freeway}), the observed state $s_t$ is an image. Here, we use a Gaussian distribution for the state marginal estimation. Using the same notation as above, the sufficient statistics for the Gaussian distribution are given by empirical mean and variance $\mu^{h,w,c}_t = \frac{\sum_{t'=0}^t s_{t'}^{h,w,c}}{t}$, $\sigma^{h,w,c}_t = \frac{\sum_{t'=0}^t (\mu^{h,w,c}_t-s_{t'}^{h,w,c})^2}{t} $. The sufficient statistic for the entire state marginal distribution is then given by $\theta_t = \{\mu_t^{h,w,c}, \sigma_t^{h,w,c}: h\in H, w\in W, c \in C\}$.

\newpage
\section{Additional Experiments}
\label{sec:app-butterfly-density}
Here we present additional results on the impact of butterfly density on the behavior of the \textbf{S-Adapt} agent.
\begin{figure}[h!]
\centering
\begin{subfigure}{.33\textwidth}
  \centering
  \includegraphics[width=.98\linewidth]{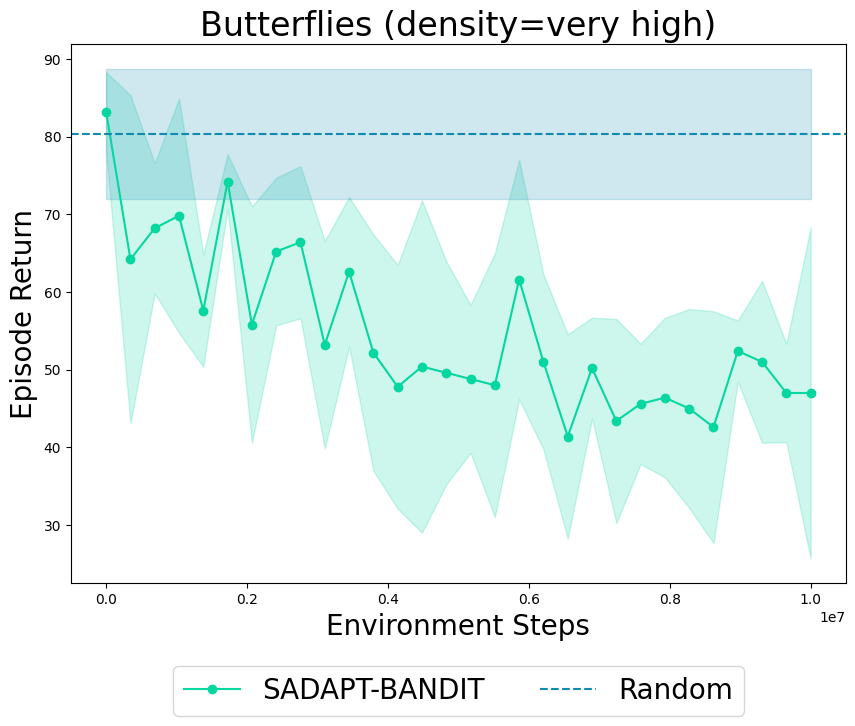}
  \caption{Very High Density}
  \label{fig:very_high_density}
\end{subfigure}%
\begin{subfigure}{.33\textwidth}
  \centering
  \includegraphics[width=.98\linewidth]{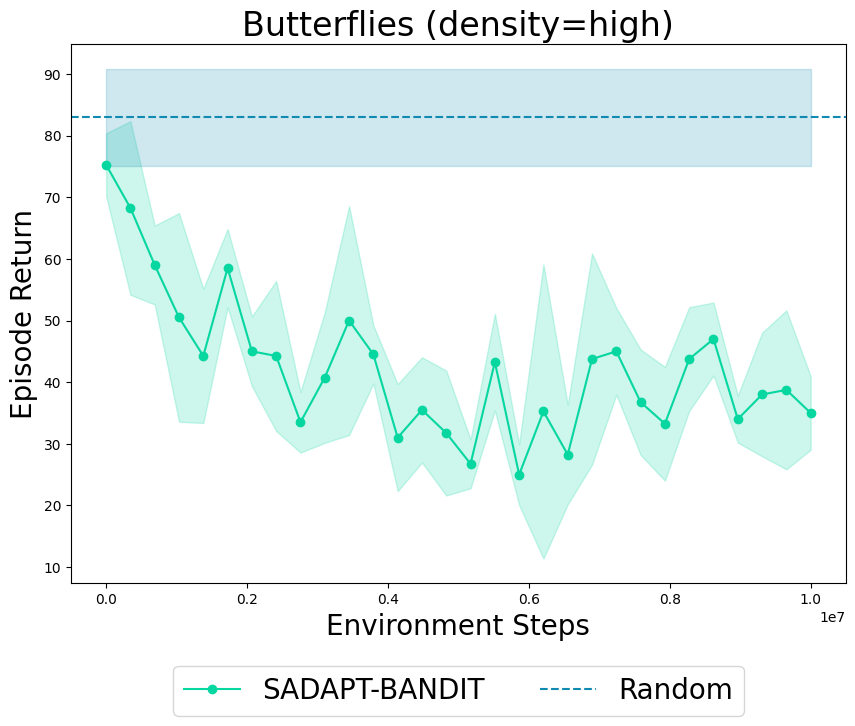}
  \caption{High Density}
  \label{fig:high_density}
\end{subfigure}
\begin{subfigure}{.33\textwidth}
  \centering
  \includegraphics[width=.98\linewidth]{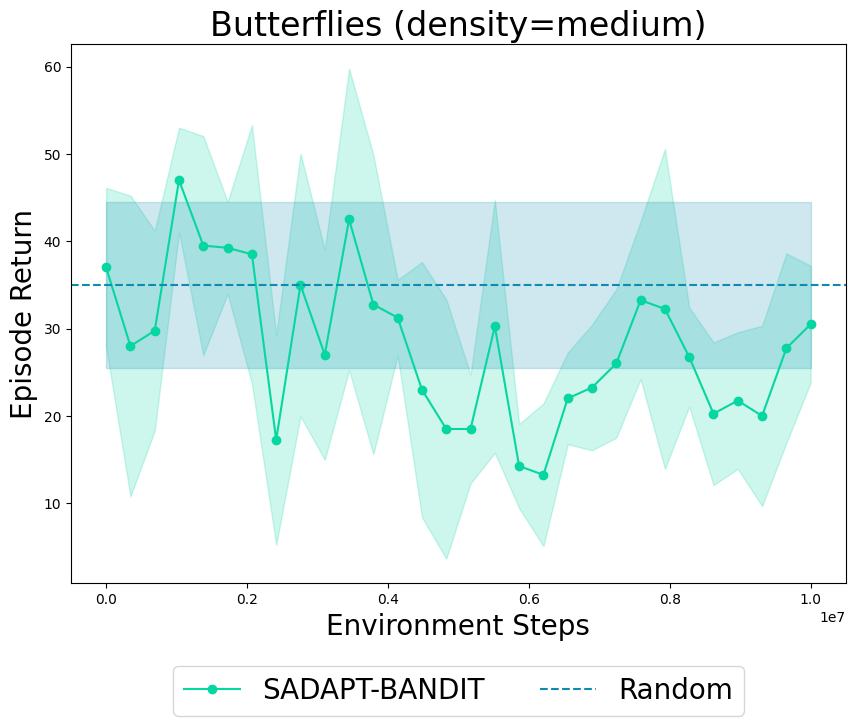}
  \caption{Medium Density}
  \label{fig:medium_density}
\end{subfigure}
\newline
\begin{subfigure}{.33\textwidth}
  \centering
  \includegraphics[width=.98\linewidth]{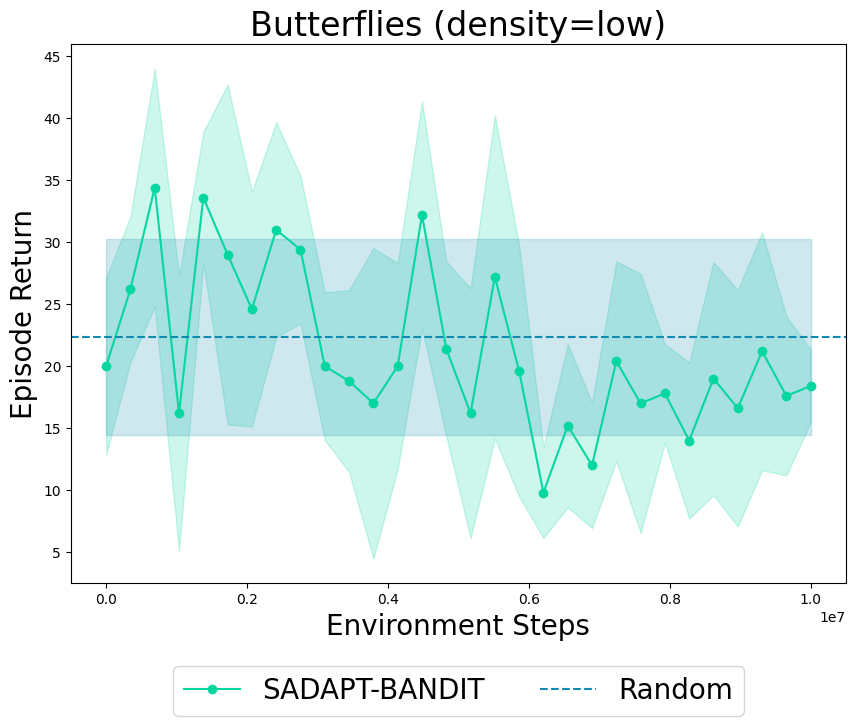}
  \caption{Low Density}
  \label{fig:low_density}
\end{subfigure}
\begin{subfigure}{.32\textwidth}
  \centering
  \includegraphics[width=.98\linewidth]{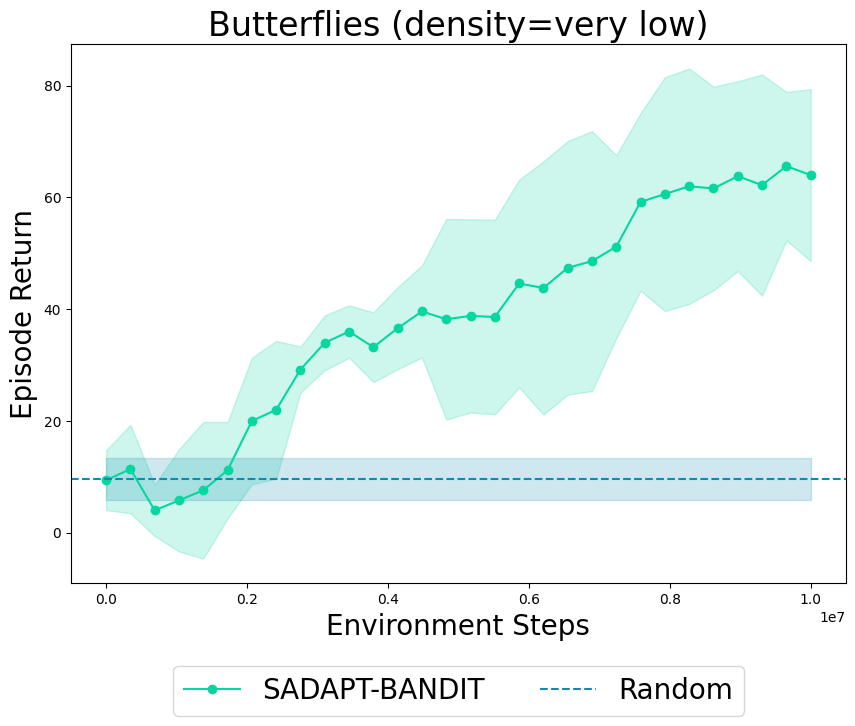}
  \caption{Very Low Density}
  \label{fig:very_low_density}
\end{subfigure}
  \caption{Average episode return of the \textbf{S-Adapt} agent (average over 5 seeds, with one shaded standard deviation) over various butterflies densities in the \textit{Butterflies} (large) environment. At (very) high density (\Cref{fig:very_high_density,fig:high_density}), the \textbf{Random} agent resembles the \textbf{S-Min} agent and catches large number of butterflies as indicated by the high episode return. Hence, the \textbf{S-Adapt} agent converges to surprise-maximization to induce large absolute difference in entropy from the \textbf{Random} agent and avoids butterflies as indicated by the low episodic return. In contrast, at very low density (\Cref{fig:very_low_density}), the \textbf{Random} agent is unable to catch butterflies and resembles the \textbf{S-Max} agent. The \textbf{S-Adapt} agent converges to surprise-minimization and almost catches all the butterflies as indicated by the high episodic return. At medium and low densities (\Cref{fig:medium_density,fig:low_density}), the \textbf{S-Adapt} agent oscillates between surprise-maximization and surprise-minimization as they roughly induce the same absolute difference in entropy.}
  \label{fig:buetterflies_density}
\end{figure}

\end{document}